\documentclass[letterpaper]{article} 
\usepackage{aaai25}  
\usepackage{times}  
\usepackage{helvet}  
\usepackage{courier}  
\usepackage[hyphens]{url}  
\usepackage{graphicx} 
\usepackage[numbers]{natbib}
\usepackage{caption} 
\usepackage{algorithm}
\usepackage{algorithmic}
\usepackage{hyperref}

\usepackage{newfloat}
\usepackage{listings}
\DeclareCaptionStyle{ruled}{labelfont=normalfont,labelsep=colon,strut=off} 
\floatstyle{ruled}
\newfloat{listing}{tb}{lst}{}
\floatname{listing}{Listing}
\usepackage{booktabs}
\usepackage[accsupp]{axessibility}  
\usepackage{enumerate}
\usepackage{amsmath}
\usepackage{url}
\usepackage{multirow}
\usepackage{pifont}
\usepackage{subcaption}
\usepackage{makecell}
\usepackage{exscale}

\usepackage{tabularx}
\usepackage{adjustbox}
\usepackage{tcolorbox}
\usepackage{colortbl}
\usepackage{soul}

\usepackage{tikz}
\tikzstyle{every picture}+=[remember picture]
\tikzstyle{na} = [shape=rectangle,inner sep=0pt]

\title{Thinking Racial Bias in Fair Forgery Detection: Models, Datasets and Evaluations}
\author{
    Decheng Liu\textsuperscript{\rm 1}\equalcontrib,
    Zongqi Wang\textsuperscript{\rm 2}\equalcontrib,
    Chunlei Peng\textsuperscript{\rm 1}\thanks{Corresponding author.}
    Nannan Wang\textsuperscript{\rm 1},
    Ruimin Hu\textsuperscript{\rm 1},
    Xinbo Gao\textsuperscript{\rm 3}
}
\affiliations{
    \textsuperscript{\rm 1}Xidian University, Xi’an, China\\ \textsuperscript{\rm 2}Tsinghua University, Beijing, China\\
    \textsuperscript{\rm 3}Chongqing University of Posts and Telecommunications, Chongqing, China
    
    dchliu@xidian.edu.cn, zq-wang24@mails.tsinghua.edu.cn, clpeng@xidian.edu.cn, nnwang@xidian.edu.cn, 
    hrm1964@163.com, gaoxb@cqupt.edu.cn
    
}

\begin{document}

\maketitle

\begin{abstract}
Due to the successful development of deep image generation technology, forgery detection plays a more important role in social and economic security. Racial bias has not been explored thoroughly in the deep forgery detection field. In the paper, we first contribute a dedicated dataset called the Fair Forgery Detection (FairFD) dataset, where we prove the racial bias of public state-of-the-art (SOTA) methods. 
Different from existing forgery detection datasets, the self-constructed FairFD dataset contains a balanced racial ratio and diverse forgery generation images with the largest-scale subjects. Additionally, we identify the problems with naive fairness metrics when benchmarking forgery detection models. 
To comprehensively evaluate fairness, we design novel metrics including Approach Averaged Metric and Utility Regularized Metric, which can avoid deceptive results. 
We also present an effective and robust post-processing technique, Bias Pruning with Fair Activations (BPFA), which improves fairness without requiring retraining or weight updates. 
Extensive experiments conducted with 12 representative forgery detection models demonstrate the value of the proposed dataset and the reasonability of the designed fairness metrics. 
By applying the BPFA to the existing fairest detector, we achieve a new SOTA. 
Furthermore, we conduct more in-depth analyses to offer more insights to inspire researchers in the community.
The source code is available at 
\emph{https://github.com/liudan193/Fairness-Benchmark-for-Face-Forgery-Detection}.
\end{abstract}

\section{Introduction}
\label{sec:intro}


Face forgery refers to the creation of fake images or videos of a person’s face using conventional techniques or deep learning methods. These forgeries can be used to spread misinformation, commit fraud, or even blackmail people. There are numerous methods are proposed for detecting face forgery \cite{rossler2019faceforensics++, afchar2018mesonet, wang2020cnn, tan2019efficientnet, nguyen2019capsule, li1811exposing, li2020face, dang2020detection, ni2022core, cao2022end, yan2023ucf, qian2020thinking, liu2021spatial, luo2021generalizing}. 
Although an increasing number of advanced face forgery detection technologies are being developed, the racial fairness of these detectors is consistently overlooked by researchers~\cite{masood2023deepfakes}. 
Detectors with severe racial bias can lead to significant social impact. These detectors might disproportionately label faces from a particular racial group as fake, thereby indicating discrimination towards this particular racial group. Therefore, when a detector is ready for deployment, evaluating and analyzing its fairness is a crucial process. 
However, although there is extensive available research about the fairness in machine learning to draw upon~\cite{agarwal2018reductions, agarwal2019fair, hardt2016equality, tian2024fairseg, shen2024finetuning}, evaluating the fairness in face forgery detection systems remains difficult. This is due to several distinct differences between face forgery detection and other deep learning tasks. 

This work aims to fill the gap in research on racial fairness in face forgery detection by proposing an accurate, comprehensive and credible fairness evaluation system. To achieve this goal, we analyze the shortcomings of existing evaluation components (i.e. dataset and metric), and our corresponding solutions. 
\textbf{(1) Dataset.} Existing face forgery detection datasets have a limited number of subjects. We find performance fluctuations significantly across subjects, so individual fairness may overshadow group fairness, which will make the evaluation results inaccurate. 
It also can be found that different forgery approaches have different fairness levels, limited forgery approaches will lead to a non-comprehensive result. 
Otherwise, undefined ethnicity (faces from two ethnicities are swapped) will lead to an inaccurate result. 
\textbf{(2) Metric.} We also propose two issues (Bias Offset and Aggregation Distortion) that will cause deceptive results. Bias Offset arises because existing fairness metrics typically use overall average accuracy for calculations instead of assessing each forgery method separately. This way may obscure some biases as different forgery methods may have different privileged races. 
Aggregation Distortion arises because detectors often show significant utility variations across different forgery techniques. Even if two forgery methods exhibit the same bias, detectors with lower utility can be more unfair. Treating each forgery method as equal will lead to unreliable results. 

\begin{figure*}[t!]
\centering
    \includegraphics{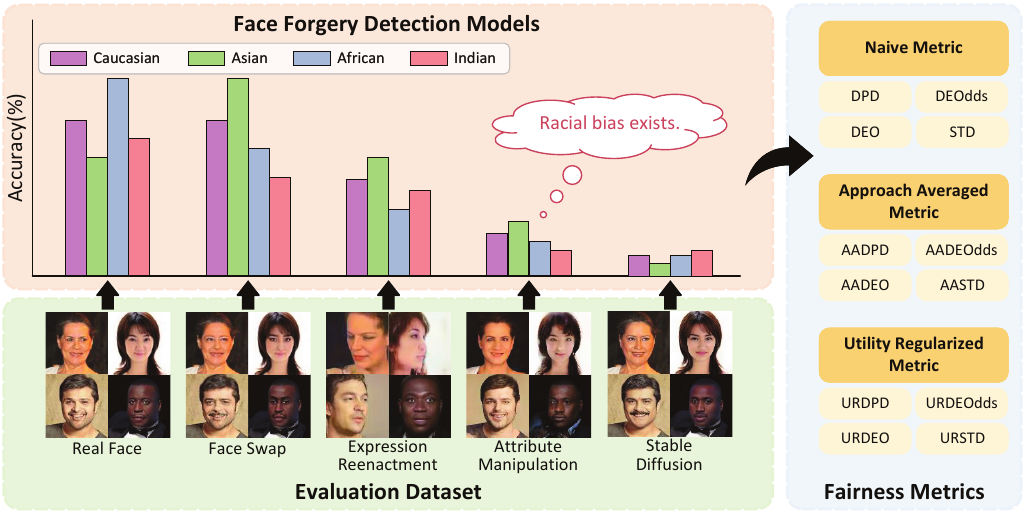}
\caption{Workflow of fairness evaluation in forgery detection. We first construct an evaluation dataset containing a large number of subjects, diverse forgery approaches, and racial balance. Subsequently, we obtain the test results of the forgery detector on each race and forgery method. Finally, we comprehensively evaluate the detector using three sets of 12 fairness metrics in total. }
\label{fig:main}
\end{figure*}

To tackle these problems, we firstly introduce the Fair Forgery Detection (FairFD) dataset for racial bias evaluation, which contains the largest scale subjects, and incorporates diverse forgery approaches including \emph{Face Swapping:} FaceSwap~\cite{FaceswapProject}, SimSwap~\cite{chen2020simswap}, \emph{Expression Reenactment:} FastReen~\cite{zakharov2020fast}, DualReen~\cite{hsu2022dual}, \emph{Face Editing:} MaskGAN~\cite{lee2020MaskGAN}, StarGAN~\cite{choi2018stargan}, StyleGAN~\cite{karras2019style}, \emph{Diffusion-Based:} SDSwap~\cite{DiffusionFaceswap}, DCFace~\cite{kim2023dcface}, Face2Diffusion~\cite{shiohara2024face2diffusion} and \emph{Transformer-Based:} FSRT~\cite{rochow2024fsrt}. 
And the self-constructed FairFD dataset does not have any undefined ethnicity annotations. 
For the specific metric, we address the mentioned two issues by introducing the Approach Averaged Metric, which calculates fairness separately for each forgery approach and then aggregates them, and the Utility Regularized Metric, which uses the utility to regularize the fairness. 

In addition to our evaluation system, we also introduce a new pruning approach called BPFA (Bias Pruning with Fair Activations). 
The designed BPFA leverages an innovative pruning metric to identify weights with the least impact on model utility while contributing most significantly to bias (e.g., racial bias). 
By pruning these weights, BPFA successfully enhances fairness without compromising utility. 
As a post-processing method, BPFA enhances fairness without any retraining. And it can be applied to any detector, including those already trained with existing fairness learning strategies, to further improve fairness performance. 
The workflow of fairness evaluation is illustrated in Figure~\ref{fig:main}.
Sufficient experimental results prove that the proposed BPFA is an efficient, plug-and-play and robust pruning scheme, outperforming other baseline pruning methods by a significant margin. 

The key contributions are summarized as follows: 

\begin{itemize}
    \item[$\bullet$] 
    To our knowledge, it is the early exploration to introduce a comprehensive racial bias evaluation benchmark for forgery detection, providing a large-scale dataset, fairness metrics, and evaluation protocols. We newly introduce the Fair Forgery Detection (FairFD) dataset for racial bias in forgery detection evaluation, which contains the largest scale subjects, race-balanced ratio and incorporates diverse forgery approaches.

    \item[$\bullet$] 
    We identify the bias offset and aggregation distortion problems with naive fairness metrics. Following, the novel Approach Averaged Metric and Utility Regularized Metric are designed to address the mentioned issues. Extensive experimental results demonstrate the limited fairness of existing SOTA methods and validate the value of our proposed metric. 

    \item[$\bullet$] 
    We propose the Bias Pruning with Fair Activations (BPFA) method to improve the fairness of forgery detectors. Extensive experiments demonstrate the advantages of our method, particularly its efficiency, plug-and-play nature and robustness. By combining BPFA with the fairest detector, we achieve a new SOTA in racial fairness performance. Specifically, we offer in-depth analyses and insightful observations to advance the community. 
\end{itemize}

\begin{table*}[t!]
\centering
\setlength{\tabcolsep}{1mm}
\small
\begin{tabular}{@{}cccccccccccc@{}}
    \toprule
    \multirow{2}{*}{\textbf{Dataset}} & \multicolumn{5}{c}{\textbf{Race Rate}}                                                & \multirow{2}{*}{\textbf{\begin{tabular}[c]{@{}c@{}}Race\\ Balance\end{tabular}}} & \multirow{2}{*}{\textbf{\begin{tabular}[c]{@{}c@{}}Undefined\\ Ethnicity\end{tabular}}} & \multirow{2}{*}{\textbf{\begin{tabular}[c]{@{}c@{}}Subject\\ Number\end{tabular}}} & \multirow{2}{*}{\textbf{\begin{tabular}[c]{@{}c@{}}App- \\ roach\end{tabular}}} & \multirow{2}{*}{\textbf{\begin{tabular}[c]{@{}c@{}}Real Img\\ Number\end{tabular}}} & \multirow{2}{*}{\textbf{\begin{tabular}[c]{@{}c@{}}Fake Img\\ Number\end{tabular}}} \\ \cmidrule(lr){2-6}
                                      & \textbf{Caucasian} & \textbf{Asian} & \textbf{Indian} & \textbf{African} & \textbf{Others} &                                                                                   &                                                                                         &                                                                                       &                                      &                                                                                       &                                                                                       \\ \midrule
    FF++~\cite{rossler2019faceforensics++}                              & $\sim$43.9\%       & $\sim$16.8\%   & $\sim$3.2\%     & $\sim$3.8\%      & $\sim$32.3\%    & \ding{56}                                                        & \textit{Yes}                                                              & $\sim$1000                                                                            & 1                                    & 73k                                                                               & 266k                                                                              \\
    UADFV~\cite{yang2019exposing}                             & 97.96\%            & 2.04\%         & 0               & 0                & 0               & \ding{56}                                                        & \textit{Yes}                                                              & 49                                                                                    & 1                                    & 241                                                                                   & 252                                                                                   \\
    CelebDF-v2~\cite{li2020celeb}                        & 88.10\%            & 5.10\%         & 0               & 6.80\%           & 0               & \ding{56}                                                        & \textit{Yes}                                                              & 59                                                                                    & 1                                    & 225k                                                                                & 2,116k                                                                              \\
    DFDC~\cite{dolhansky2020deepfake}                              & -                  & -              & -               & -                & -               & \ding{56}                                                        & \textit{Yes}                                                              & 960                                                                                   & 8                                    & 488k                                                                                & 1,783k                                                                              \\
    DF-1.0~\cite{jiang2020deeperforensics}                            & $\sim$25\%         & $\sim$25\%     & $\sim$25\%      & $\sim$25\%       & 0               & \ding{51}                                                        & \textit{Yes}                                                              & 100                                                                                   & 1                                    & \multicolumn{2}{c}{total 17,600k}                                                                                                                                             \\
    ForgeryNet~\cite{he2021forgerynet}                        & -                  & -              & -               & -                & -               & \ding{56}                                                        & \textit{Yes}                                                              & 5400                                                                                  & 15                                   & 1438k                                                                             & 1457k                                                                             \\
    \textbf{FairFD(ours)}                              & $\sim$25\%         & $\sim$25\%     & $\sim$25\%      & $\sim$25\%       & 0               & \ding{51}                                                        & \textit{No}                                                              & 11430                                                                                 & 11                                    & 52k                                                                               & 572k                                                                              \\ \bottomrule
\end{tabular}
\caption{Face Forgery Detection Dataset Comparison. Our dataset is race-balanced, with no undefined races, the maximized number of subjects and forgery approaches exhibit diversity. }
\label{tab:dataset_comparison}
\end{table*}

\section{Related Work}

\subsection{Fairness in Face Forgery Detection} 

\noindent \textbf{Fairness Algorithm.} Fairness in face forgery detection is a relatively novel topic. DAG-FDD~\cite{ju2024improving} is first proposed to address fairness without demographic information by setting a probability threshold for minority groups to ensure low error rates for all groups meeting this threshold. DAW-FDD~\cite{ju2024improving} utilizes demographic information to design losses to ensure similar performance across specified groups. PFGDFD~\cite{lin2024preserving} improves fairness by using disentanglement loss to separate demographic and forgery features. 

\noindent \textbf{Deepfake Dataset.} We summarize the information of existing datasets in Table~\ref{tab:dataset_comparison}. We provide the proportions of each race, along with whether the datasets are race-balanced. Additionally, we present whether undefined ethnicity faces are included(,i.e., faces from one race are replaced with another race. We also supply the number of subjects, the number of forgery approaches, and the total number of frames. DAG(W)-FDD~\cite{ju2024improving} and PFGDFD~\cite{lin2024preserving} directly use several of the datasets mentioned above as test data. Our work reveals inherent limitations when using these datasets for evaluation. There is currently no suitable dataset to evaluate the fairness of forgery detection. We also give details of widely used face forgery datasets in the section "Face Forgery Detection Datasets" in \textit{Supp}. 

\noindent \textbf{Fairness Metric.} To evaluate racial fairness, what we require is group fairness metrics. There are various group fairness metrics, and the selection of a metric depends on the application context. 
In this work, we consider the commonly used metrics including DPD~\cite{agarwal2018reductions, agarwal2019fair}, DEOdds~\cite{agarwal2018reductions}, DEO~\cite{hardt2016equality}, STD~\cite{wang2019racial, robinson2020face, gong2020jointly, yu2020fair, wang2023mixfairface} and our proposed novel fairness metrics. The definitions of these metrics can be found in the section "Existing Fairness Metrics" in \textit{Supp}.



\section{FairFD Dataset}

\subsection{Limitations of Current Datasets}
\label{sec:disadvantages_dataset}


\noindent \textbf{Limited Number of Subjects. } The construction process of the existing face forgery detection datasets involves collecting videos, subsequently creating forgeries, and then extracting frames. As a result, there are typically a small number of subjects and each subject has a large number of frames in these datasets. The limited number of subjects makes it challenging to draw meaningful comparisons across groups. 
Furthermore, we conducted the verification experiment to analyze and prove it in the section "Number of Subjects is Limited" in \textit{Supp}. 

\noindent \textbf{Lack of Diversity of Forgery Approaches. } In Table~\ref{tab:dataset_comparison}, only DFDC and ForgeryNet employ a variety of forgery techniques. However, we find that different forgery methods have different fairness levels. We validate this point in the subsequent Figure~\ref{fig:detailed_fairness}. Thus, we should strive to diversify forgery methods as much as possible, enabling a more comprehensive evaluation of the system's fairness. 

\noindent \textbf{Undefined Attribute Annotation. } For these identity-replaced forgery approaches, there is a possibility of faces from one ethnicity being replaced with those from another. In related work ~\cite{xu2022comprehensive}, this phenomenon is referred to as "undefined attribute annotation." Undefined attributes can also significantly lower the quality of the evaluation of racial fairness, which is ignored in existing face forgery detection datasets.


%
\subsection{FairFD Description}

Considering the mentioned limitations in existing datasets, we introduce our dataset, FairFD, aiming to address these shortcomings. FairFD endeavors to overcome previous challenges and provide a more accurate, reliable and comprehensive benchmark for evaluating fairness in face forgery detection. 
Representative examples of ours are presented in Figure~\ref{fig:main}. The overview of FairFD can be shown in the Table~\ref{tab:dataset_comparison}. Subsequently, we delve into several pivotal facets of our dataset. 

Our dataset is an image-level dataset, and for each image, there are 11 kinds of corresponding forgery images, i.e., \emph{Face Swapping:} FaceSwap~\cite{FaceswapProject}, SimSwap~\cite{chen2020simswap}, \emph{Expression Reenactment:} FastReen~\cite{zakharov2020fast}, DualReen~\cite{hsu2022dual}, \emph{Face Editing:} StarGAN~\cite{choi2018stargan}, StyleGAN~\cite{karras2019style}, MaskGAN~\cite{lee2020MaskGAN}, \emph{Diffusion-Based:} SDSwap~\cite{DiffusionFaceswap}, DCFace~\cite{kim2023dcface}, Face2Diffusion~\cite{shiohara2024face2diffusion} and \emph{Transformer-Based:} FSRT~\cite{rochow2024fsrt}. 
In addition to the forgery approach label, our approach also includes labels for four ethnicities (i.e., Caucasian, Asian, African, and Indian). Each ethnicity contains approximately 3000 subjects. 

\subsection{Source Data Collection and Forgery Process}

To align with our requirements, which include having racial labels, and containing a sufficient number of subjects, we use the RFW~\cite{wang2019racial} dataset as pristine images. The RFW dataset comprises face images with four racial labels (,i.e., Caucasian, Asian, African, and Indian), containing approximately 3000 subjects for each racial group, with a roughly equal distribution. Each subject has approximately $3\sim7$ images. All images in the RFW dataset have a resolution of $400\times400$ pixels. Besides, the images are carefully selected to maintain similar distributions in terms of age, gender, yaw angle, and pitch angle. See details in "Detailed Distribution of FairFD" in \textit{Supp}.
%
To reduce the human resources, we use RFW as the source data. 

To achieve the goal of diversity, we choose various approaches and techniques. We classify the forgery methods into face swap, expression reenactment, attribute manipulation and advanced forgery methods(stable diffusion and transformer). See details in the section "Classification of Forgery Approaches" in \textit{Supp}. We reimplement these methods and apply them to the source data. Details configuration and process of forgery crafting can be found in the section "Forgery Crafting Process" in \textit{Supp}. 

\section{The Proposed Evaluation Metrics}

Even though we have obtained a reliable dataset for evaluating deepfake detection's fairness, we still can not get a credible evaluation result due to existing fairness metrics having two flaws. Firstly, the \textit{bias offset} may lead to an underestimation of racial bias. Secondly, \textit{aggregation distortion} can result in biased evaluation outcomes favoring specific forgery approaches. Below, we introduce the two flaws and their corresponding solutions respectively. 

We make corrections to four widely used metrics: DPD~\cite{agarwal2018reductions}, DEOdds~\cite{agarwal2018reductions}, DEO~\cite{hardt2016equality}, and STD~\cite{wang2019racial}. For clarity, we leverage DPD to introduce our new metric in the following discussions as an example. 
The following is the definition of DPD: 

\begin{equation}
    DPD=\max_{s,s'\in \mathbb{S},s\ne s'}\left| P\left( \hat{Y}\,\,|\,\,S=s \right) -P\left( \hat{Y}\,\,|\,\,S=s' \right) \right|,
\end{equation}

\noindent where $\hat{Y}$ is the predicted labels. $\mathbb{S}$ represents the set of sensitive attributes, ${s\in \mathbb{S}}$ and $\mathbb{S}=\{\text{Caucasian}, \text{Asian}, \text{Indian}, \text{African}\}$. 

\subsection{Bias Offset Problem } 
\label{sec:bias_offset}

\begin{figure}
    \centering
    \includegraphics{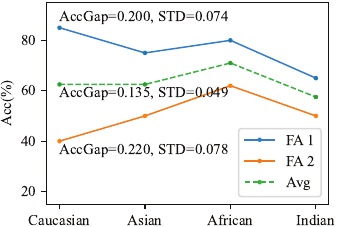}
    \caption{A face forgery detection model exhibits different biases for different forgery approaches. 
    }
\label{fig:metric_diff_fairness_level}
\end{figure}

Bias offset refers to bias that will be partially obscured due to the calculation process of existing fairness metrics. 
Existing fairness metrics do not calculate separately for each forgery method instead of the final averaged performance scores. 
However, this way may obscure certain biases, which we call bias offset. 
Taking an example, in Figure~\ref{fig:metric_diff_fairness_level}, face forgery detectors may exhibit different biases for various forgery approaches. 
We calculate AccGap (the maximum differences in accuracy) and STD (standard deviation) for each forgery approach. 
In this example, both Forgery Approach 1 (FA1) and Forgery Approach 2 (FA2) exhibit an AccGap greater than 0.2 and an STD greater than 0.07. However, for Forgery Approach 1(FA1), the performance of Caucasians is better than Asians, while for FA2, the performance of Asians is better than Caucasians. In this situation, when calculating the fairness score using the final averaged performance scores, bias is to some extent offset, resulting in a smaller bias score. 
AccGap is less than 0.2, and the STD is less than 0.07 calculated using average accuracy. We refer to this phenomenon as \emph{bias offset}. 
A more reliable way is to calculate fairness metrics separately for each forgery method and average them. We call this novel strategy as \textit{Approach Averaged Metric}. 



\begin{alignat}{2}
\text{AADPD} & = \frac{1}{|\mathbb{F}|} \sum_{f\in \mathbb{F}}{\max_{s,s'\in \mathbb{S},s\ne s'\vphantom{i}}} \nonumber \bigl| P\left( \hat{Y}\,|\,S=s,F=f \right) \nonumber \\
& - P\left( \hat{Y}\,|\,S=s',F=f \right) \bigr|,
\label{eq:method_averaged_metric}
\end{alignat}

\noindent where $\mathbb{F}$ donates real face and forgery approaches. $f\in \mathbb{F}$ and $\mathbb{F}$ is the set of forgery methods. 

\subsection{Aggregation Distortion Problem } 

\label{sec:aggregation_distortion}

Various forgery approaches not only exhibit different fairness situations but also demonstrate distinct levels of performance. For example, in Figure.~\ref{fig:detailed_utility} and Figure.~\ref{fig:detailed_fairness}, this is clearly evident. We identify the aggregation distortion problem where even if we calculate fairness scores separately for each forgery method and average them together, the averaged result can achieve a distorted fairness score due to the performance difference. 

Directly averaging fairness scores when employing common fairness metrics might lead to misleading conclusions. For instance, consider two approaches with the accuracy of 20\% and 80\% respectively. We assume that both approaches yield a bias of 10\% if we employ DEO as the fairness metric. Then, we calculate a simple average, which is also 10\%. This would lead us to focus solely on the absolute differences in error rates without taking into account the variations in baselines. If the racial biases are both calculated to be 10\%, the forgery method with only a 20\% accuracy would evidently be much more unfair. This oversimplified average fails to capture the substantial disparity in performance between the two methods. 
To address this issue, we propose a fixed version. For each forgery approach, we have: 



\begin{alignat}{2}
\text{URDPD} & = \frac{1}{|\mathbb{F}|} \sum_{f\in \mathbb{F}}{\max_{s,s'\in \mathbb{S},s\ne s'\vphantom{i}}} \nonumber \bigl| P\left( \hat{Y}\,|\,S=s,F=f \right) \nonumber \\
& - P\left( \hat{Y}\,|\,S=s',F=f \right) \bigr| \enspace / \enspace ACC_{F=f},
\label{eq:performance_regularized_metric}
\end{alignat}

\noindent where $ACC_{F=f}$ calculates the accuracy of given different forgery approaches. 

After applying Eq.~\ref{eq:performance_regularized_metric} to each forgery method, we calculate their results and then obtain the final fairness score by averaging them. We refer to this approach as \textit{Utility Regularized Metric}. This nuanced method acknowledges the significance of each forgery approach, providing a more accurate and insightful evaluation of fairness in the context of the diverse fairness and performance landscape. 
In summary, we recommend not relying on a single fairness metric but rather considering a combination of multiple metrics to collectively reflect the fairness of a detector.

\section{Bias Pruning with Fair Activations}

In this section, we present the Bias Pruning with Fair Activations (BPFA) approach, which develops a novel pruning metric combining weights and the fairness of activations to determine weight importance. Then, we prune those with the lowest pruning scores based on a predefined pruning rate by layer.
Here we utilize an unstructured pruning strategy. 
Noting that the proposed BPFA can be directly extended to process other biases except for racial bias.

\noindent \textbf{Pruning Metric.} Consider a convolutional layer weights $W$ of shape $(C_{out}, C_{in}, S_{Ker}^{h}, S_{Ker}^{w})$, where $C_{out}$ represents the number of output filters, and each filter has dimension $(C_{in}, S_{Ker}^{h}, S_{Ker}^{w})$. For one data sample, the output of this layer is denoted as $X$ with shape $(C_{out}, S_{out}^{h}, S_{out}^{w})$. The L2 norm by filter of $X$ is represented as $\lVert X \rVert _2\in \mathbb{R}^{C_{out}}$. We compute the average L2 norm across all samples from a specific race ${s\in\mathbb{S}}$, denoted by $Z^{s} = \lVert X \rVert _2^{s}$. 
For each filter, we then calculate the standard deviation of these norms across all races, which serves as the bias for that filter: 

\begin{equation}
BIAS_{i} = std\left( \left\{ Z_{i}^{s} \right\}_{s\in\mathbb{S}} \right),
\end{equation}

\noindent where $std( \cdot )$ denotes the standard deviation. This bias measures the variability of outputs across different races. The pruning score (PS) for each weight $W_{ijkm}$ in the convolutional layer at the position ${(i,j,k,m)}$ is then calculated by combining the weight with respect to the computed bias: 

\begin{equation}
PS_{ijkm} = \frac{\left| W_{ijkm} \right|}{BIAS_{i}}.
\end{equation}

By comparing the pruning scores, we can identify and potentially remove weights that have the least impact on model utility but the greatest impact on bias. This allows us to reduce bias and improve fairness without sacrificing performance. Note that while our method is illustrated using convolutional layers as an example, it can be easily extended to linear layers.

\section{Benchmark Experiments}

\subsection{Experimental Setup}
\label{sec:exper_setup}

\noindent \textbf{Dataset.} We use FF++ (c23) as our training set. Specifically, for each video, we select 32 frames, crop the facial region, and finally resize it to 256 × 256. We utilize the preprocessed data provided by~\cite{yan2023deepfakebench}, which has already undergone the aforementioned operations. 
Our proposed new dataset serves as the testing set. As our dataset inherently consists of face images with backgrounds and bodies removed, there is no need for additional face cropping. Subsequently, we resize the images to 256 × 256 for inference. Note that we still provide the original dataset with a resolution of 400 × 400 for scenarios requiring higher resolution. 

\noindent \textbf{Algorithms.} We summarize the face forgery detection algorithms in the section "Face Forgery Detection Algorithms Categories" in \textit{Supp}. For a comprehensive and fair analysis, we select several representative algorithms. For spatial-based detectors, we choose Xception~\cite{rossler2019faceforensics++}, RECCE~\cite{cao2022end}, UCF~\cite{yan2023ucf}, Capsule~\cite{nguyen2019capsule}, FFD~\cite{dang2020detection} and CORE~\cite{ni2022core}. For frequency-based detectors, we select F3Net~\cite{qian2020thinking}, SPSL~\cite{liu2021spatial} and SRM~\cite{luo2021generalizing}. For fairness-enhanced detectors, we select DAG~\cite{ju2024improving}(Xception as base model), DAW~\cite{ju2024improving}(Xception as base model) and PFGDFD~\cite{lin2024preserving}(UCF as base model). 
%
In detail, these models are trained with the Adam optimization algorithm with a learning rate of 0.0002 and an epoch number of 10. The batch size is 32. And data augmentation methods including image compression, horizontal flip and rotation are applied. 
However, when applying these data augmentation methods to DAG, we find that its fairness level significantly deteriorated. For a fair comparison, we report below the results using data augmentation. Meanwhile, the results without data augmentation are presented in the section "Results without Data Augmentation" in \textit{Supp}.

\begin{table*}[t!]
\centering
\setlength{\tabcolsep}{1mm}
\small
\begin{tabular}{cccccccccccccc}
\toprule
\multicolumn{2}{c}{\multirow{2}{*}{Fairness Metric}}                                                   & \multicolumn{6}{c}{Spatial-based}                      & \multicolumn{3}{c}{Frequency-based}  & \multicolumn{3}{c}{Fairness-enhanced}  \\ \cmidrule(r){3-8} \cmidrule(r){9-11} \cmidrule(r){12-14}
\multicolumn{2}{c}{}                                                                                   & Xception & RECCE  & UCF    & Capsule & FFD    & CORE   & F3Net  & SPSL   & SRM    & \small{DAG}    & \small{DAW}    &\small{PFGDFD}     \\ \midrule \midrule 
\multirow{4}{*}{\begin{tabular}[c]{@{}c@{}}Naive\\ Metric\end{tabular}}                     & DPD$\downarrow$      & 0.1810   & 0.1338 & 0.1765 & 0.0969  & 0.1099 & 0.0951 & \textbf{0.0674} & \underline{\textbf{0.0203}} & 0.0990 & 0.1723 & \underline{\textbf{\textit{0.0513}}} & \textbf{\textit{0.0805}}     \\
                                                                                            & DEOdds$\downarrow$   & 0.1666   & 0.1264 & 0.1495 & 0.0902  & 0.1005 & 0.0798 & \textbf{\textit{0.0763}} & \underline{\textbf{0.0304}} & \textbf{0.0714} & 0.2288 & \underline{\textbf{\textit{0.0593}}} & 0.1396    \\
                                                                                            & DEO$\downarrow$      & 0.2088   & 0.1548 & 0.2014 & 0.1118  & 0.1242 & 0.1084 & \textbf{0.080}1 & \underline{\textbf{0.0215}} & 0.1090 & 0.2105 & \underline{\textbf{\textit{0.0611}}} & \textbf{\textit{0.1032}}    \\
                                                                                            & STD$\downarrow$      & 0.0647   & 0.0474 & 0.0631 & 0.0343  & 0.0398 & 0.0342 & \textbf{0.0265} & \underline{\textbf{0.0080}} & 0.0355 & 0.0636 & \underline{\textbf{\textit{0.0195}}} & \textbf{\textit{0.0328}}    \\ \midrule
\multirow{4}{*}{\begin{tabular}[c]{@{}c@{}}Approach\\ Averaged\\ Metric\end{tabular}}       & AADPD$\downarrow$    & 0.2024   & 0.1572 & 0.2175 & 0.1323  & 0.1552 & \textbf{0.1147} & \textbf{\textit{0.1158}} & \underline{\textbf{0.0556}} & 0.1413 & 0.2201 & \underline{\textbf{\textit{0.0735}}} & 0.1393    \\
                                                                                            & AADEOdds$\downarrow$ & 0.1669   & 0.1302 & 0.1630 & 0.1034  & 0.1196 & \textbf{0.0858} & 0.0961 & \underline{\textbf{0.0481}} & \textbf{\textit{0.0925}} & 0.2324 & \underline{\textbf{\textit{0.0662}}} & 0.1560    \\
                                                                                            & AADEO$\downarrow$    & 0.2095   & 0.1626 & 0.2284 & 0.1381  & 0.1623 & \textbf{\textit{0.1205}} & \textbf{0.1197} & \underline{\textbf{0.0571}} & 0.1511 & 0.2177 & \underline{\textbf{\textit{0.0749}}} & 0.1360    \\
                                                                                            & AASTD$\downarrow$    & 0.0750   & 0.0578 & 0.0809 & 0.0493  & 0.0576 & \textbf{\textit{0.0449}} & \textbf{0.0448} & \underline{\textbf{0.0219}} & 0.0531 & 0.0834 & \underline{\textbf{\textit{0.0283}}} & 0.0530    \\ \midrule
\multirow{4}{*}{\begin{tabular}[c]{@{}c@{}}Utility\\ Regularized\\ Metric\end{tabular}}     & URDPD$\downarrow$    & 0.1357   & 0.1118 & 0.1523 & 0.0808  & 0.1037 & \textbf{0.0803} & \textbf{\textit{0.0806}} & \underline{\textbf{0.0320}} & 0.0904 & 0.1474 & \underline{\textbf{\textit{0.0555}}} & 0.0881    \\
                                                                                            & URDEOdds$\downarrow$ & 0.1057   & 0.0852 & 0.1069 & 0.0639  & 0.0763 & \textbf{0.0567} & 0.0625 & \underline{\textbf{0.0299}} & \textbf{\textit{0.0584}} & 0.1445 & \underline{\textbf{\textit{0.0440}}} & 0.0986    \\
                                                                                            & URDEO$\downarrow$    & 0.1417   & 0.1171 & 0.1614 & \textbf{0.0842}  & 0.1092 & \textbf{\textit{0.0850}} & \textbf{0.0842} & \underline{\textbf{0.0324}} & 0.0968 & 0.1480 & \underline{\textbf{\textit{0.0578}}} & 0.0860    \\
                                                                                            & URSTD$\downarrow$    & 0.0501   & 0.0410 & 0.0565 & \textbf{0.0301}  & 0.0384 & 0.0313 & \textbf{\textit{0.0312}} & \underline{\textbf{0.0126}} & 0.0339 & 0.0559 & \underline{\textbf{\textit{0.0214}}} & 0.0335    \\ \midrule
\multirow{1}{*}{\begin{tabular}[c]{@{}c@{}}Utility\end{tabular}}                            & AUC$\uparrow$      & \textbf{\textit{0.6911}}   & 0.6897 & \underline{\textbf{\textit{0.7214}}} & 0.6815  & \underline{\textbf{0.7304}} & 0.6864 & 0.6564 & 0.6763 & \textbf{0.7102} & 0.6672 & 0.6604 & 0.6302     \\ \bottomrule										
\end{tabular}
\caption{Bias evaluation on FairFD for 12 face forgery detectors using Naive Metrics, Approach Averaged Metrics, Utility Regularized Metrics. For each row, the best values are \underline{\textbf{underlined and bolded}}, followed by the second-best values which are \underline{\textbf{\textit{underlined, bolded, and italicized}}}, the third-best values are \textbf{bolded}, and the fourth-best values are \textbf{\textit{bolded and italicized}}. }
\label{tab:benchmark}
\end{table*}

\begin{figure*}[h!]
\centering
    \includegraphics{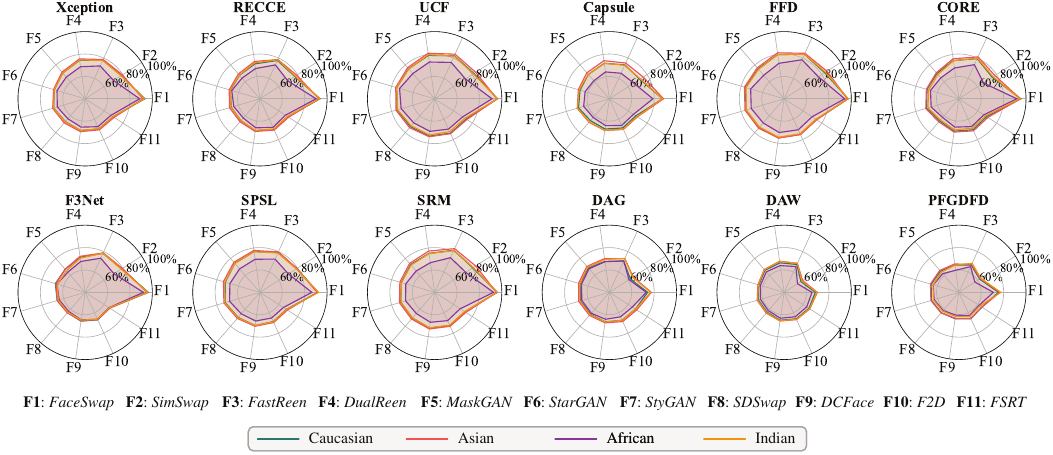}
\caption{Detailed utility (AUC) for diverse races, forgery approaches, detectors. }
\label{fig:detailed_utility}
\end{figure*}

\subsection{Benchmarking Fairness of Face Forgery Detectors}

\begin{table*}[!t]
\centering
\setlength{\tabcolsep}{0.9mm}
\small
\begin{tabular}{cc|cccc|cccc|cccc|cc}
\hline
\multicolumn{2}{c|}{}                                                               & \multicolumn{4}{c|}{Naive Metric$\downarrow$}                                                                                                                         \rule{0pt}{2.5ex}& \multicolumn{4}{c|}{Approach Averaged Metric$\downarrow$}                                                                                                               & \multicolumn{4}{c|}{Utility Regularized Metric$\downarrow$}                                                                                                       & \multicolumn{2}{c}{Utility$\uparrow$}                                             \\ \cline{3-16} 
\multicolumn{2}{c|}{\multirow{-2}{*}{Method}}                                       & DPD                                     & DEOdds                                  & DEO                                     & STD                                     & \begin{tabular}[c]{@{}c@{}}MA\\ DPD\end{tabular}                                   & \begin{tabular}[c]{@{}c@{}}MA\\ DEOdds\end{tabular}                                & \begin{tabular}[c]{@{}c@{}}MA\\ DEO\end{tabular}                                   & \begin{tabular}[c]{@{}c@{}}MA\\ STD\end{tabular}                                   & \begin{tabular}[c]{@{}c@{}}UR\\ DPD\end{tabular}                                   & \begin{tabular}[c]{@{}c@{}}UR\\ DEOdds\end{tabular}                                & \begin{tabular}[c]{@{}c@{}}UR\\ DEO\end{tabular}                                   & \begin{tabular}[c]{@{}c@{}}UR\\ STD\end{tabular}                                   & AUC                                     & ACC                                     \\ \hline
\multicolumn{1}{c|}{}                         \rule{0pt}{2.5ex}& Original                            & 0.0203                                  & 0.0304                                  & 0.0215                                  & 0.0080                                  & 0.0556                                  & 0.0481                                  & 0.0571                                  & 0.0219                                  & 0.0320                                  & 0.0299                                  & 0.0324                                  & 0.0126                                  & 0.6763                                  & 0.7618                                  \\ \cline{2-16} 
\multicolumn{1}{c|}{}                         \rule{0pt}{2.5ex}& WEIG                         & 0.0183                                  & 0.0258                                  & 0.0201                                  & \textbf{0.0072}                         & 0.0564                                  & 0.0451                                  & 0.0586                                  & 0.0219                                  & 0.0324                                  & 0.0277                                  & 0.0334                                  & 0.0126                                  & 0.6769                                  & 0.7615                                  \\ \cline{2-16} 
\multicolumn{1}{c|}{}                         \rule{0pt}{2.5ex}& RoBA                         & 0.1128                                  & 0.1598                                  & 0.1395                                  & 0.0445                                  & 0.1462                                  & 0.1616                                  & 0.1432                                  & 0.0583                                  & 0.0893                                  & 0.1024                                  & 0.0867                                  & 0.0356                                  & 0.6331                                  & 0.7037                                  \\ \cline{2-16} 
\multicolumn{1}{c|}{\multirow{-4}{*}{SPSL}}   \rule{0pt}{2.5ex}& \cellcolor[HTML]{E7E6E6}BPFA   & \cellcolor[HTML]{E7E6E6}\textbf{0.0181} & \cellcolor[HTML]{E7E6E6}\textbf{0.0209} & \cellcolor[HTML]{E7E6E6}\textbf{0.0200} & \cellcolor[HTML]{E7E6E6}\textbf{0.0072} & \cellcolor[HTML]{E7E6E6}\textbf{0.0473} & \cellcolor[HTML]{E7E6E6}\textbf{0.0357} & \cellcolor[HTML]{E7E6E6}\textbf{0.0496} & \cellcolor[HTML]{E7E6E6}\textbf{0.0182} & \cellcolor[HTML]{E7E6E6}\textbf{0.0265} & \cellcolor[HTML]{E7E6E6}\textbf{0.0218} & \cellcolor[HTML]{E7E6E6}\textbf{0.0275} & \cellcolor[HTML]{E7E6E6}\textbf{0.0102} & \cellcolor[HTML]{E7E6E6}\textbf{0.6862} & \cellcolor[HTML]{E7E6E6}\textbf{0.8055} \\ \hline
\multicolumn{1}{c|}{}                         \rule{0pt}{2.5ex}& Original                            & 0.1099                                  & 0.1005                                  & 0.1242                                  & 0.0398                                  & 0.1552                                  & 0.1196                                  & 0.1623                                  & 0.0576                                  & 0.1037                                  & 0.0763                                  & 0.1092                                  & 0.0384                                  & 0.7304                                  & 0.5751                                  \\ \cline{2-16} 
\multicolumn{1}{c|}{}                         \rule{0pt}{2.5ex}& WEIG                         & 0.1098                                  & 0.1003                                  & 0.1240                                  & 0.0398                                  & 0.1550                                  & 0.1194                                  & 0.1621                                  & 0.0576                                  & 0.1035                                  & 0.0761                                  & 0.1090                                  & 0.0384                                  & 0.7304                                  & 0.5751                                  \\ \cline{2-16} 
\multicolumn{1}{c|}{}                         \rule{0pt}{2.5ex}& RoBA                                & -                                  & -                                  & -                                  & -                                  & -                                  & -                                  & -                                  & -                                  & -                                  & -                                  & -                                  & -                                  & 0.5967                                  & -                                  \\ \cline{2-16} 
\multicolumn{1}{c|}{\multirow{-4}{*}{FFD}}    \rule{0pt}{2.5ex}& \cellcolor[HTML]{E7E6E6}BPFA   & \cellcolor[HTML]{E7E6E6}\textbf{0.1096} & \cellcolor[HTML]{E7E6E6}\textbf{0.0999} & \cellcolor[HTML]{E7E6E6}\textbf{0.1237} & \cellcolor[HTML]{E7E6E6}\textbf{0.0397} & \cellcolor[HTML]{E7E6E6}\textbf{0.1546} & \cellcolor[HTML]{E7E6E6}\textbf{0.1189} & \cellcolor[HTML]{E7E6E6}\textbf{0.1617} & \cellcolor[HTML]{E7E6E6}\textbf{0.0574} & \cellcolor[HTML]{E7E6E6}\textbf{0.1032} & \cellcolor[HTML]{E7E6E6}\textbf{0.0758} & \cellcolor[HTML]{E7E6E6}\textbf{0.1087} & \cellcolor[HTML]{E7E6E6}\textbf{0.0382} & \cellcolor[HTML]{E7E6E6}\textbf{0.7305} & \cellcolor[HTML]{E7E6E6}\textbf{0.5760} \\ \hline
\multicolumn{1}{c|}{}                         \rule{0pt}{2.5ex}& Original                            & 0.0805                                  & 0.1396                                  & 0.1032                                  & 0.0328                                  & 0.1393                                  & 0.1560                                  & 0.1360                                  & 0.0530                                  & 0.0881                                  & 0.0986                                  & 0.0860                                  & 0.0335                                  & 0.6302                                  & 0.6019                                  \\ \cline{2-16} 
\multicolumn{1}{c|}{}                         \rule{0pt}{2.5ex}& WEIG                         & 0.0789                                  & 0.1340                                  & 0.1012                                  & 0.0319                                  & 0.1349                                  & 0.1494                                  & 0.1320                                  & 0.0513                                  & 0.0853                                  & \textbf{0.0944}                         & 0.0835                                  & 0.0324                                  & 0.6298                                  & 0.6021                                  \\ \cline{2-16} 
\multicolumn{1}{c|}{}                         \rule{0pt}{2.5ex}& RoBA                                & -                                  & -                                  & -                                  & -                                  & -                                  & -                                  & -                                  & -                                  & -                                  & -                                  & -                                  & -                                  & 0.5468                                  & -                                  \\ \cline{2-16} 
\multicolumn{1}{c|}{\multirow{-4}{*}{\begin{tabular}[c]{@{}c@{}}PFG- \\ DFD\end{tabular}}} \rule{0pt}{2.5ex}& \cellcolor[HTML]{E7E6E6}BPFA & \cellcolor[HTML]{E7E6E6}\textbf{0.0594} & \cellcolor[HTML]{E7E6E6}\textbf{0.1337} & \cellcolor[HTML]{E7E6E6}\textbf{0.0796} & \cellcolor[HTML]{E7E6E6}\textbf{0.0238} & \cellcolor[HTML]{E7E6E6}\textbf{0.1079} & \cellcolor[HTML]{E7E6E6}\textbf{0.1442} & \cellcolor[HTML]{E7E6E6}\textbf{0.1006} & \cellcolor[HTML]{E7E6E6}\textbf{0.0411} & \cellcolor[HTML]{E7E6E6}\textbf{0.0644} & \cellcolor[HTML]{E7E6E6}0.0969          & \cellcolor[HTML]{E7E6E6}\textbf{0.0578} & \cellcolor[HTML]{E7E6E6}\textbf{0.0245} & \cellcolor[HTML]{E7E6E6}\textbf{0.6445} & \cellcolor[HTML]{E7E6E6}\textbf{0.7415} \\ \hline
\end{tabular}
\caption{Experiments with different fairness pruning methods. We highlight the best method for each metric in \textbf{bold}. And we use '-' to indicate methods that cause severe performance degradation, rendering the detector unusable even setting a pruning rate as low as 0.1\%. }
\label{tab:pruning_result}
\end{table*}

\textbf{Benchmark Results.} The benchmark results (shown in Table~\ref{tab:benchmark}) present a comprehensive evaluation of 12 face forgery detectors using various fairness metrics. We highlight the four fairest detectors using different colors. Based on the results, we draw the following significant observations: 
\emph{(1) Current face forgery detectors all exhibit a high degree of racial bias.} 
The DPD metric of SPSL can achieve 0.0203 with the smallest racial bias.
This indicates that the difference in the probability of classifying faces as fake between the most advantaged and disadvantaged groups is 2.03\%. 
When separately calculating and averaging for each forgery method, the AADPD metric reaches 5.56\%. 
On the other hand, for the least fair detector UCF, the difference in the probability of classifying faces as fake between the most advantaged and disadvantaged groups is 17.65\%. This reminds researchers to address the racial bias in existing face forgery detection models. 
\emph{(2)~Current face forgery detectors have racial bias variation.} Comparing the least fair detector UCF with the most fair detector SPSL, the former's URDPD is 4.76 times that of the latter, URDEOdds is 3.58 times, URDEO 4.98 times, and URSTD is 4.48 times, showing significant gap in racial bias. 
Other detectors also exhibit varying degrees of racial bias. 
Furthermore, we observe that three frequency-based detectors, SPSL, F3Net, and SRM, consistently demonstrate a smaller racial bias across all fairness metrics. We conduct an in-depth investigation into this in the section "Analyses and Discussions" in \textit{Supp}. 

\noindent \textbf{Detailed Utility Results.} To present more detailed results, we present the AUC for each detector, each race, and each forgery method in Figure~\ref{fig:detailed_utility}. Results show that different forgery methods exhibit varying levels of utility. This validates the advantage of Utility Regularized Metric. 

\begin{figure}[!t]
\centering
    \includegraphics{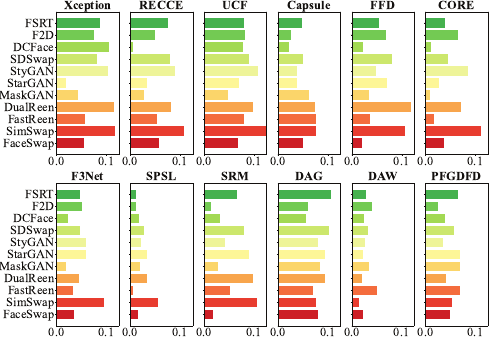}
\caption{Fairness (STD) for different forgery approaches. }
\label{fig:detailed_fairness}
\end{figure}

\noindent \textbf{Detailed Fairness Results.} We present the standard deviation (STD) of the ACC for the four races in Figure~\ref{fig:detailed_fairness} for each forgery method(including Real Face). Our findings reveal that different forgery methods exhibit varying levels of fairness, and different detectors rank the fairness of these forgery methods differently. This validates the advantage of the proposed Approach Averaged Metric. 

\section{Evaluating BPFA}

\subsubsection{Baseline Algorithm}

We select two baseline methods for comparison. The first WEIG uses only the absolute values of the weights as the pruning score, and the second RoBA uses only the reciprocal of the bias of the activations. 
More details about these baselines are shown in \textit{Supp}.
For all three methods, we prune the parameters with the lowest pruning scores. 
These pruning baselines can be directly applied in existing SOTA forgery detection models.

\subsubsection{Results Analysis}

The experimental results under optimal pruning rates for each detector and method are shown in Table~\ref{tab:pruning_result}. 
%
It can be found that the proposed method BPFA consistently outperforms all baseline methods, achieving superior fairness without compromising utility.
Although WEIG generally preserves good utility and enhances fairness, its improvements in fairness are not as significant as those achieved by BPFA. On the other hand, RoBA exhibits highly unstable performance. 
It results in improving fairness but at the cost of significantly reduced utility, which can render the model nearly unusable in the forgery detection task. 
These results prove the superior performance of BPFA in enhancing both utility and fairness for forgery detection.
\emph{It is encouraging to find that SPSL+BPFA achieves the new state-of-the-art performance}.
The only parameter in our method is the pruning rate. The ablation study for pruning rate is detailed in three tables in the section "Ablation Study on Pruning Rate" in \textit{Supp}. 
Our findings indicate that different detectors and different methods have significantly different optimal pruning rates (which correspond to the least decrease in utility (or even improvement) while achieving the best average value across the 12 fairness metrics. 

\subsection{Analyses and Discussions}

Here we conduct deeper analyses and give some insights:
(1) We train detectors using balanced training data, finding that while racial bias can be reduced, the cost of collecting balanced data is substantial;
(2) We set the optimal classification threshold for each race and then use the resulting accuracy values to calculate fairness. The conclusions show that this method is highly cost-effective and significantly improves both utility and fairness simultaneously;
(3) We analyze that frequency-based detectors exhibit superior fairness performance because of not utilizing race-sensitive information, e.g. color.
More analyses and details are shown in \textit{Supp}.
\section{Conclusion}

This paper early explores a comprehensive racial bias evaluation benchmark for forgery detection, which provides a newly self-construct dataset, fairness metric and unified protocols. 
We identify numerous disadvantages in existing datasets and fairness metrics, then propose a novel dataset FairFD dataset and two sets of fairness metrics to address these mentioned issues. 
Besides, we also propose a novel Bias Pruning with Fair Activations algorithm to improve the fairness performance without an extra training process.
Emphatically, we evaluate the fairness of multiple existing face forgery detectors. 
The results indicate the racial bias in current detectors is generally high and prove the advantages of our proposed BPFA. 
Further analyses reveal some interesting insights into the emergence of racial bias. 
%
We hope the proposed benchmark can inspire more researchers to develop the field.
In the future, we will explore a unified fairness metric for diverse biases in more kinds of datasets, and construct the video-level forgery detection datasets for more real applications.
See social impact in \textit{Supp}.

\section{Acknowledgments}

This work was supported in part by the National Natural Science Foundation of China under Grant 62306227, Grant 62276198, Grant U22A2035, Grants U22A2096, Grant 62441601 and Grant 62036007;  
in part by the Fundamental Research Funds for the Central Universities under Grant ZYTS24142, Grant QTZX23083 and Grant QTZX23042; 
in part by the Key Research and Development Program of Shaanxi (Program No. 2023-YBGY-231); in part by Young Elite Scientists Sponsorship Program by CAST under Grant 2022QNRC001; in part by the Guangxi Natural Science Foundation Program under Grant 2021GXNSFDA075011; in part by the Shaanxi Province Core Technology Research and Development Project under grant 2024QY2-GJHX-11; in part by Open Research Project of Key Laboratory of Artificial Intelligence Ministry of Education under Grant AI202401, in part by the Nanning Scientific Research and Technological Development Project 20231042; in part by the ‘111 Center’ (B16037).

\bibliographystyle{unsrt}
\bibliography{aaai25}


\section{Full Related Works} 

\subsection{Other Work on Fairness in Face Forgery Detection} 

The research on fairness in face forgery detection is still relatively limited and waits for further exploration. A preliminary study~\cite{trinh2021examination} investigates bias in three commonly used face forgery detectors. They created real face images by sampling from the RFW~\cite{wang2019racial} and UTKFace~\cite{zhifei2017cvpr}. Next, they generate fake faces by blending two faces. However, they do not explicitly consider any fairness metrics. This simple study lacks a reasonable evaluation system but still verifies face forgery detectors have a significant racial bias to some extent. 
Study in~\cite{xu2022comprehensive} annotates five popular deepfake detection datasets with age, gender, ethnicity, etc. Due to the racial imbalance of current datasets, they also propose a metric to deal with the unbalanced test dataset. 
Another study~\cite{nadimpalli2022gbdf} create a gender-balanced dataset (GBDF) sampled from the FF++, Celeb-DF, and DF-1.0. This approach involves a limited number of subjects, which restricts the dataset's capability for fairness evaluations. 

\subsection{Classification of Forgery Approaches} 
\label{sec:classification_approaches}

There are several approaches for creating fake faces. Here, we provide a brief overview of the classification of forgery methods: 

\textbf{1. Identity-replaced Forgery Approach} refers to substituting the original identity in an image, e.g., FaceSwap. 
\textbf{FaceSwap}~\cite{FaceswapProject, chen2020simswap, perov2020deepfacelab, yu2022migrating} involves replacing the face of a person in a video or image with another person’s face. The method usually uses deep learning algorithms to detect and extract the faces of the two people and then swaps them. 

\textbf{2. Identity-remained Forgery Approach} retains the original identity and alters other facial attributes, e.g., Attribute Manipulation and Expression Reenactment. 
\textbf{Attribute Manipulation}~\cite{shen2017learning, choi2018stargan, lee2020MaskGAN, sun2022fenerf, xu2022transeditor} involves manipulating the attributes of a face, such as skin color, gender, and eye shape, while preserving the identity of the person. The method usually adopts a GAN. 
\textbf{Expression Reenactment}~\cite{thies2015real, thies2016face2face, wu2018reenactgan, zakharov2020fast, hsu2022dual} involves transferring the facial expressions of one person to another person’s face. The method usually detect and track the facial landmarks of the two people and then transfers the expression from the source face to the target face. 

\subsection{Existing Fairness Metrics} 
\label{sec:existing_metrics}

Commonly, researchers use the performance metric difference between privileged groups and unprivileged groups as a fairness metric. Demographic Parity Difference (DPD)~\cite{agarwal2018reductions, agarwal2019fair} utilizes the difference in positive rate, which represents the proportion of data predicted to be positive, as its fairness metric. The difference in Equalized Odds (DEOdds)~\cite{agarwal2018reductions} utilizes the average of the differences in true positive rate and false positive rate as its fairness metric. The difference in Equal Opportunity (DEO)~\cite{hardt2016equality} utilizes the difference in true positive rate solely as a fairness metric. 
In face recognition scenarios, the Standard Deviation (STD) of performance metrics across different groups is often used~\cite{wang2019racial, robinson2020face, gong2020jointly, yu2020fair, wang2023mixfairface}. Equity-Scaled Segmentation Performance (ESSP) proposes to evaluate segmentation performance and group fairness simultaneously in medical image segmentation scenarios~\cite{tian2024fairseg}. In the context of the generative model, the frequency of each group in the generated images is computed, and the average difference in frequency between each pair of groups is calculated as fairness metric~\cite{shen2024finetuning}. Moreover, there are numerous works proposing more suitable metrics based on the application scenarios.

Due to the complexity of fairness evaluation, numerous fairness metrics are proposed from various perspectives to cater to different scenarios. Therefore, we need to leverage multiple fairness metrics to evaluate the fairness in face forgery detection comprehensively. Here, we present the four most commonly used fairness metrics and outline their respective applicable scenarios. 

\noindent \textbf{Demographic Parity Difference (DPD):} In the context of face forgery detection, where label 1 represents fake face, DPD reflects the model's inclination to categorize faces of a specific race as fake. When people perceive the classification of faces as fake as a form of discrimination, we can leverage DPD to assess the extent of bias in the model. 


\begin{equation}
    DPD=\max_{s,s'\in \mathbb{S},s\ne s'}\left| P\left( \hat{Y}\,\,|\,\,S=s \right) -P\left( \hat{Y}\,\,|\,\,S=s' \right) \right|, 
\end{equation}

\noindent where $\hat{Y}$ is the predicted labels. $\mathbb{S}$ represents the set of sensitive attributes, $S \in \mathbb{S}$ and $\mathbb{S}=\{\text{Caucasian}, \text{Asian}, \text{Indian}, \text{African}\}$. 

\noindent \textbf{Difference in Equalized Odds (DEOdds):} DPD may fail in certain situations~\cite{grant2023equalized}. Considering a scenario where a face forgery detector classifies faces of African and Caucasian individuals as fake at a similar rate, but the model makes different types of errors for the two groups. Specifically, for African faces, the false positive rate is significantly higher than for Caucasian faces, while for Caucasian individuals, the false negative rate is higher. In such a case, we can use DEOdds. 


\begin{tiny}
\begin{equation}
    DEOdds=\frac{1}{2}\underset{y=\left\{ 0,1 \right\}}{\Sigma}\max_{s,s'\in \mathbb{S},s\ne s'}\left| P\left( \hat{Y}|Y=y,S=s \right) -P\left( \hat{Y}|Y=y,S=s' \right) \right|.
\end{equation}
\end{tiny}


\noindent \textbf{Equal Opportunity (DEO):} DEO has more relaxed conditions compared to DEOdds. When assessing the fairness between group A and group B, only the images of faces that are inherently fake need to be considered. DEO solely focuses on true positive rates and does not capture the overall classification differences. 


\begin{scriptsize}
\begin{equation}
DEO=\max_{s,s'\in \mathbb{S},s\ne s'}\left| P\left( \hat{Y}|Y=1,S=s \right) -P\left( \hat{Y}|Y=1,S=s' \right) \right|.
\end{equation}
\end{scriptsize}


\noindent \textbf{Standard Deviation (STD):} STD differs from metrics mentioned above. STD considers the overall variability rather than just the differences between the best and worst-performing ethnicities. We use accuracy as the performance metric. 


\begin{equation}
STD=std\left( \left\{ acc\left( S=s \right) _{s\in \mathbb{S}} \right\} \right) ,
\end{equation}
where $std$ is a function that calculates the standard deviation of given list. $acc$ calculates the accuracy of a given race. 


%
\subsection{Face Forgery Detection Datasets} 
\label{sec:dataset_deepfake}

Below we give simple description of some widely used face forgery datasets. 

\noindent \textbf{FaceForensics++(FF++)}~\cite{rossler2019faceforensics++} is a forensics dataset that consists of 1000 original video sequences downloaded from the Internet(,i.e.,YouTube). The videos are manipulated with four automated face manipulation methods. For face swap, they use Deepfakes~\cite{DeepfakesProject} and FaceSwap~\cite{FaceswapProject}. For expression reenactment, they use Face2Face~\cite{thies2016face2face} and NeuralTextures~\cite{thies2019deferred}. 

\noindent \textbf{UADFV}~\cite{yang2019exposing} contains videos of varying classes, with each video being classified as either real or fake. The dataset is relatively small, with only 98 videos, but it has been found to be convenient in terms of how the data is formatted. 

\noindent \textbf{CelebDF-v2}~\cite{li2020celeb} is a comprehensive collection for deepfake forensics, comprising 590 real videos and 5,639 DeepFake videos featuring celebrities with high-quality. These videos were generated through an enhanced synthesis process, ensuring superior visual quality and better representing the DeepFake content prevalent on the internet. 

\noindent \textbf{Deepfake Detection Challenge(DFDC)}~\cite{dolhansky2020deepfake} is created by Facebook in partnership with other industry leaders and academic experts. The dataset consists of 128,154 videos featuring 960 paid actors. The dataset was used in a Kaggle competition to create new and better models to detect manipulated media. 

\noindent \textbf{DeeperForensics-1.0(DF-1.0)}~\cite{jiang2020deeperforensics} contains 60,000 videos and 17.6 million frames with 100 consented actors. The actors in DF-1.0 have four skin tones: white, black, yellow, brown, with roughly balanced ratio. Different from other datasets, DF-1.0 attach importance to high-quality and diversity of source face videos. Each actor has various poses, expressions, and illuminations. And DF-1.0 has more real videos than fake videos with a ratio of 5:1. 

\noindent \textbf{ForgeryNet}~\cite{he2021forgerynet} is a very large dataset for real-world face forgery detection, with 2.9 million images, 221,247 videos and 15 forgery approaches. And a pipeline of conducting various face forgery approaches are proposed.

\subsection{Face Forgery Detection Algorithms Categories} 
\label{sec:deepfake_detectors}

%
%
Current face forgery detectors can be roughly divided into two categories: Spatial-based, and Frequency-based. 
\textbf{Spatial-based method}~\cite{rossler2019faceforensics++, afchar2018mesonet, wang2020cnn, tan2019efficientnet, nguyen2019capsule, li1811exposing, li2020face, dang2020detection, ni2022core, cao2022end, yan2023ucf} is based on the spatial domain features(,i.e., forgery clues) of the fake face. Such as RECCE~\cite{cao2022end} utilizes an encoder to reconstruct real face images, thereby exploring the differences between real and fake images in the spatial domain. The differences in the reconstruction of the two are then used as guidance to train a classifier for detecting deepfakes. 
\textbf{Frequency-based method}~\cite{qian2020thinking, liu2021spatial, luo2021generalizing} identifies distinctions between real and fake faces in frequency domain, which are used to detect whether the face is forged. Such as F3Net~\cite{qian2020thinking} utilizes frequency-aware decomposed image components and local frequency statistics to explore forgery patterns, enabling effective forgery detection. 

\subsection{Fairness in Face Recognition} 
\label{sec:fairness_in_face_recognition}

An increasing number of researchers are are now paying attention to societal issues of artificial intelligence, including adversarial example~\cite{chen2024content, wang2024apd, franzmeyer2024illusory}, privacy protection~\cite{xu2023federated, qiao2024offline, pittaluga2023ldp}, and fairness concerns. 

The research on fairness in face forgery detection is similar to the study of fairness in face recognition. 
RFW~\cite{wang2019racial} is first introduced as a race balanced test dataset for face recognition. They also propose IMAN which uses a deep information maximization adaptation network to align global distribution to decrease race gap at domain-level, and learns the discriminative target representations at cluster level. 
Another balanced face recognition dataset BFW~\cite{robinson2020face} is proposed to evaluate the fairness of face recognition system both for gender and ethnic groups and this work also shows variations in the optimal scoring threshold for face-pairs across different subgroups. 

The issue of racial bias in machine learning has garnered significant attention in multi fields. The research on fairness in face forgery detection is akin to the study of fairness in face recognition. 
RFW~\cite{wang2019racial} is first introduced as a balanced test dataset for ethnic group. IMAN~\cite{wang2019racial} uses a deep information maximization adaptation network to align global distribution to decrease race gap at domain-level, and learns the discriminative target representations at cluster level. 
Another balanced faces dataset BFW~\cite{robinson2020face} is proposed to evaluate the fairness of face recognition system both for gender and ethnic groups and this work also shows variations in the optimal scoring threshold for face-pairs across different subgroups. 
DebFace~\cite{gong2020jointly} uses a de-biasing adversarial network to extract disentangled feature representations for both unbiased face recognition and demographics estimation and adopts adversarial learning to minimize correlation among feature factors so as to abate bias influence. 
\cite{yu2020fair} involves multiple preprocessing methods to improve the dual-shot face detector, data re-sampling to balance the data distribution, and multiple data enhancement methods to increase accuracy performance and proposes a linear-combination strategy is adopted to benefit from multi-model fusion. 
Meta Balanced Network~\cite{wang2021meta} uses meta-learning algorithm to learn adaptive margins in large margin loss to mitigate the algorithmic bias in face recognition models. 
MixFairFace~\cite{wang2023mixfairface} proposes MixFair Adapter to determine and reduce the identity bias of training samples. 

\subsection{Model Pruning} 
\label{sec:pruning}

Current pruning methods typically use two types of signals: the magnitude~\cite{han2015learning, frankle2019lottery, blalock2020state} of the neurons and the absolute value of the neuron activations~\cite{sunsimple}. Generally, the smaller these values, the less contribution they make to the model. 

In our method, we combine the magnitude of the neurons and the bias of activations across different races as the pruning score. The resulting pruning score helps identify neurons that have a small contribution to model performance but a large contribution to racial bias.

\section{Details of Crafting Dataset}

\subsection{Number of Subjects is Limited} 
\label{sec:number_subject}

\begin{table}
    \centering
    \begin{tabular}{|l|l|l|l|l|}
    \hline
    \textbf{} & Caucasian & Asian  & African & Indian \\ \hline
    TPR   & 0.7764    & 0.7365 & 0.6289  & 0.6801 \\ \hline
    TNR   & 0.9502    & 0.9652 & 0.9314  & 0.9646 \\ \hline
    \end{tabular}
    \caption{Evaluation with TPR and TNR for each race on FF++ subset. }
    \label{tab:id_all}
\end{table}

We propose that \textbf{the limited number of subjects makes it challenging to draw meaningful comparisons across groups.} 
Here we do a verification experiment to explain and prove it. We utilize the FF++ dataset, retains only the Caucasian, Asian, African, and Indian subsets. The remaining dataset consists of a total of 677 subjects. For each race, 16 subjects are chosen as the test dataset, and the rest serve as the training dataset (9:1 approximately). We train an Xception model on the training dataset for deepfake detection and evaluate its performance on the test dataset. We follow the hyperparameters specified in \cite{yan2023deepfakebench} and train the model for 10 epochs. 

\begin{figure*}[tb]
  \centering
  \includegraphics[width=1.0\linewidth]{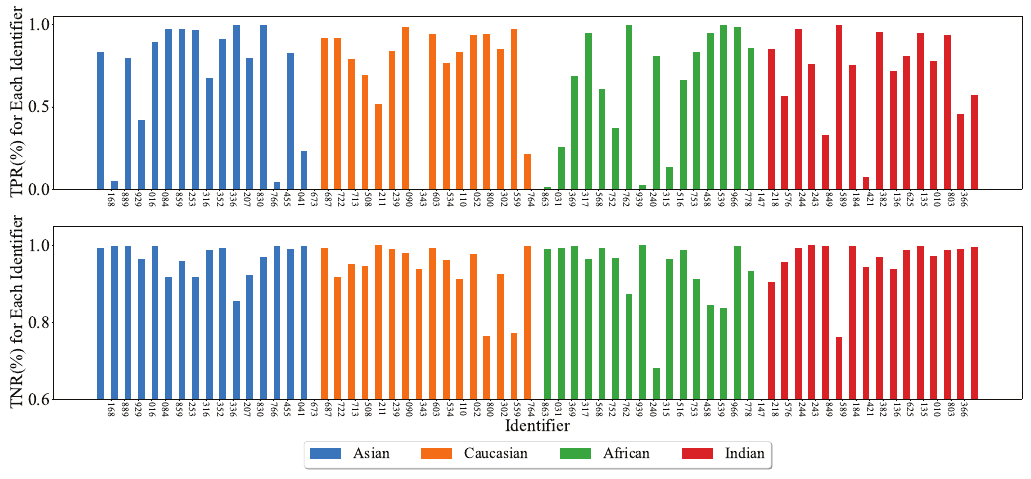}
  \caption{Evaluation with TPR and TNR for for each subject on FF++ subset. }
  \label{fig:id_each_id}
\end{figure*}

In Table~\ref{tab:id_all}, we present the TPR and TNR of different ethnicities. And we can find that model performs worse on African for both TPR and TNR. It seems that this result indicates discrimination of the detectors towards African subset, but we argue that this discrimination is not credible. 
In Figure~\ref{fig:id_each_id}, we show the TPR and TNR on each subject. We can observe significant fluctuations in the model's performance across each subject. The presence of extreme results also indicates that subjects significantly influence the model's classification. 
Therefore, when the number of subjects is insufficient, it becomes challenging to capture the overall characteristics of a group, making it difficult to draw meaningful comparisons across race groups. 

\subsection{Forgery Crafting Process}
\label{sec:Forgery_Crafting_Process}

1) For face swap, we select FaceSwap~\cite{FaceswapProject} and SimSwap~\cite{chen2020simswap}. Although both methods result in a face-swapping effect, to ensure approach diversity, we select two distinct face-swapping approaches. FaceSwap~\cite{FaceswapProject} is a graphic-based face swap method. Whereas SimSwap~\cite{chen2020simswap} is a learning-based face swap method. 
2) For expression reenactment, we select FastReen~\cite{zakharov2020fast} and DualReen~\cite{hsu2022dual}. 
Both face swap and expression reenactment methods involve the use of source and target images to transfer faces or expressions from the target image to the source image. For each subject, we first designate it as the source subject. Next, within the same ethnic group, we randomly select another subject as the target subject, ensuring that the transfer occurs only within a single ethnic group as stated in section "Limitations of Current Datasets" in \textit{Supp}. Once the source-target pairs are established, these pairs remain fixed in other identity-replaced forgery approaches. Considering that each subject may have multiple images, for each specific image, we randomly select one image from its appointed target subject for the transfer, ensuring diversity and variability in the dataset. 
%
3) For attribute manipulation, we select MaskGAN~\cite{lee2020MaskGAN}. MaskGAN~\cite{lee2020MaskGAN} provides an official GUI program that allows manual manipulation of attributes by making use of face parsing. However, as we aim to automate the face forgery process, we randomly select a subset from the nose, glasses, left eye, right eye, left eyebrow, right eyebrow, left ear, right ear, mouth, upper lip, and lower lip. We then apply random dilate and erode operations to the selected parts, enlarging or reducing the chosen regions. Missing parts are filled with skin, resulting in the final manipulated outcome. 
4) For much recent forgery methods, we select Diffusion-Based(SDSwap~\cite{DiffusionFaceswap}, DCFace~\cite{kim2023dcface}, Face2Diffusion~\cite{shiohara2024face2diffusion}) and Transformer-Based(FSRT~\cite{rochow2024fsrt}). These approaches use advanced technologies: Stable Diffusion and Transformer, resulting in more realistic generated faces. 
In summary, we select three types, a total of 11 forgery approaches, which achieve the diversity requirement. 

\subsection{Detailed Distribution of FairFD}
\label{sec:Detailed_Distribution_of_FairFD}

For the number of race categories, existing work almost all uses three or four races, and we follow previous works~\cite{wang2023mixfairface, lin2024preserving, ju2024improving} with standard practice for convenience. 

\begin{figure*}[tb]
  \centering
  \includegraphics[width=1.0\linewidth]{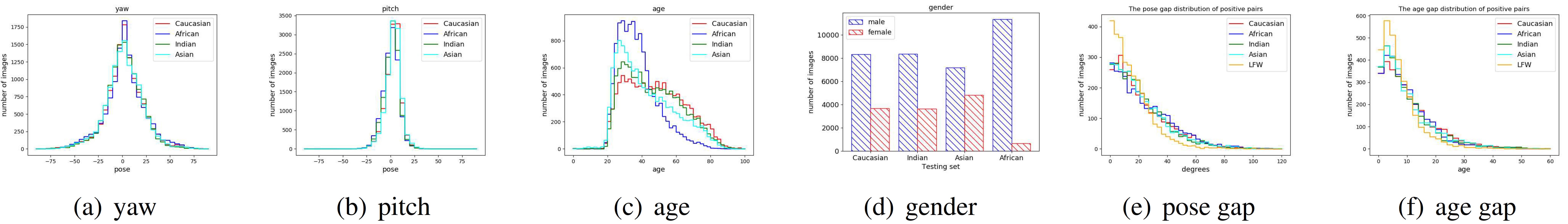}
  \caption{The distribution of other attributes (aside from racial attributes) in FairFD. This figure is adapted from RFW~\cite{wang2019racial}. }
  \label{fig:Detailed_Distribution_of_FairFD}
\end{figure*}

Moreover, to ensure an accurate evaluation, we keep distributions of other attributes (including gender, age, etc) similar across different racial groups. Since our goal is to simulate real-world racial fairness evaluation, in each race group, distributions of these attributes are simulated based on real-world distributions. These attributes (e.g., gender, age, etc.) do not introduce additional biases, as we derive our real face images from the existing, widely recognized racial bias evaluation dataset RFW~\cite{wang2019racial}. It is worth noting that these attributes are not perfectly balanced within each racial group, as we focus on 'in-the-wild' faces, aiming to simulate real-world distributions for a more meaningful practical evaluation. Refer to Figure~\ref{fig:Detailed_Distribution_of_FairFD} for detailed distributions. We ensure similar distributions of these attributes across racial groups, ensuring that our dataset does not introduce additional bias.

\section{Analyses and Discussions} 
\label{sec:analysis}

\subsubsection{Balanced Training Dataset. }

Data imbalance across races is a prevalent source of bias. To assess the impact of imbalanced training sets on racial bias, we create a dataset that balances across different racial groups. The dataset is obtained by sampling from the FF++. Because FF++ contains a very few number of African and Indian subjects, we can only select 26 subjects from each race. This balanced dataset is employed as the training set. We adopt the same data preprocessing methods and training parameters as the racially imbalanced training set. We use real face, FaceSwap, SimSwap, FastReen, DualReen and MaskGAN of FairFD as our dataset. Notably, due to the reduced dataset size, the number of epochs is doubled to $20$. 

\begin{table*}[t!]
\centering
\renewcommand{\arraystretch}{1.1}  
\tabcolsep=0.05cm 
\begin{tabular}{cccccccccc}
\toprule
\multicolumn{2}{c}{\multirow{2}{*}{Metrics}}                                                                       & \multicolumn{2}{c}{Xception}     & \multicolumn{2}{c}{F3Net}        & \multicolumn{2}{c}{RECCE}        & \multicolumn{2}{c}{UCF}          \\ \cmidrule(r){3-4} \cmidrule(r){5-6} \cmidrule(r){7-8} \cmidrule(r){9-10}
\multicolumn{2}{c}{}                                                                                                       & Unbalanced      & Balanced        & Unbalanced      & Balanced        & Unbalanced      & Balanced        & Unbalanced      & Balanced        \\ \cmidrule(r){1-2} \cmidrule(r){3-4} \cmidrule(r){5-6} \cmidrule(r){7-8} \cmidrule(r){9-10}
\multicolumn{1}{c}{\multirow{4}{*}{\begin{tabular}[c]{@{}c@{}}Naive\\ Metric\end{tabular}}}                     & DPD      & 0.1538          & \textbf{0.0852} & \textbf{0.0764} & 0.0817          & 0.1317          & \textbf{0.0836} & 0.1782          & \textbf{0.0908} \\
\multicolumn{1}{c}{}                                                                                            & DEOdds   & 0.1672          & \textbf{0.0837} & 0.0877          & \textbf{0.0630} & 0.1379          & \textbf{0.0663} & 0.1655          & \textbf{0.0674} \\
\multicolumn{1}{c}{}                                                                                            & DEO      & 0.2095          & \textbf{0.1076} & 0.1030          & \textbf{0.1027} & 0.1777          & \textbf{0.1027} & 0.2334          & \textbf{0.1104} \\
\multicolumn{1}{c}{}                                                                                            & STD      & 0.0563          & \textbf{0.0334} & \textbf{0.0278} & 0.0329          & 0.0473          & \textbf{0.0314} & 0.0635          & \textbf{0.0355} \\ \hline
\multicolumn{1}{c}{\multirow{4}{*}{\begin{tabular}[c]{@{}c@{}}Approach\\ Averaged\\ Metric\end{tabular}}}         & AADPD    & 0.1957          & \textbf{0.1030} & 0.1102          & \textbf{0.0996} & 0.1644          & \textbf{0.1044} & 0.2107          & \textbf{0.0961} \\
\multicolumn{1}{c}{}                                                                                            & AADEOdds & 0.1674          & \textbf{0.0857} & 0.0951          & \textbf{0.0691} & 0.1378          & \textbf{0.0746} & 0.1655          & \textbf{0.0674} \\
\multicolumn{1}{c}{}                                                                                            & AADEO    & 0.2099          & \textbf{0.1116} & 0.1178          & \textbf{0.1149} & 0.1777          & \textbf{0.1193} & 0.2334          & \textbf{0.1104} \\
\multicolumn{1}{c}{}                                                                                            & AASTD    & 0.0721          & \textbf{0.0412} & 0.0425          & \textbf{0.0393} & 0.0603          & \textbf{0.0405} & 0.0773          & \textbf{0.0374} \\ \hline
\multicolumn{1}{c}{\multirow{4}{*}{\begin{tabular}[c]{@{}c@{}}Utility\\ Regularized\\ Metric\end{tabular}}} & URDPD    & 0.1314          & \textbf{0.0741} & \textbf{0.0769} & 0.0770          & 0.1158          & \textbf{0.0733} & 0.1433          & \textbf{0.0723} \\
\multicolumn{1}{c}{}                                                                                            & URDEOdds & 0.1069          & \textbf{0.0573} & 0.0624          & \textbf{0.0511} & 0.0908          & \textbf{0.0505} & 0.1069          & \textbf{0.0487} \\
\multicolumn{1}{c}{}                                                                                            & URDEO    & 0.1437          & \textbf{0.0824} & \textbf{0.0842} & 0.0900          & 0.1283          & \textbf{0.0847} & 0.1614          & \textbf{0.0842} \\
\multicolumn{1}{c}{}                                                                                            & URSTD    & 0.0482          & \textbf{0.0297} & \textbf{0.0295} & 0.0302          & 0.0423          & \textbf{0.0284} & 0.0523          & \textbf{0.0281} \\ \hline
\multicolumn{1}{c}{Utility}                                                                                 & Accuracy & \textbf{0.5552} & 0.3984          & \textbf{0.5266} & 0.3308          & \textbf{0.5011} & 0.4276          & \textbf{0.5557} & 0.3682          \\ \bottomrule
\end{tabular}
\caption{Evaluations on the unbalanced and balanced training dataset. }
\label{tab:balanced_training_dataset}
\end{table*}

\begin{figure*}[t!]
  \centering
  \includegraphics[width=1.0\linewidth]{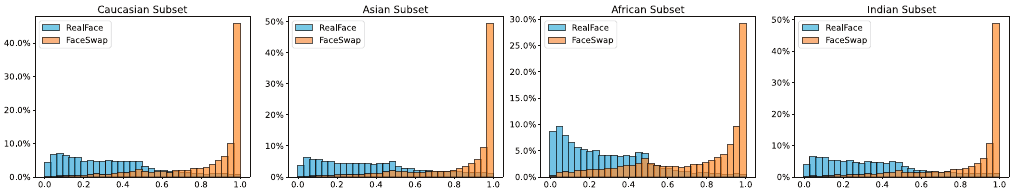}
  \caption{Probability score distribution of each race. }
  \label{fig:probability_score}
\end{figure*}

\begin{table*}
    \centering
    \renewcommand{\arraystretch}{1.2}
    \tabcolsep=0.04cm
    \begin{tabular}{ccccccc}
        \hline
        \multirow{2}{*}{Xception} & \multicolumn{4}{c}{Utility(Accuracy) ↑}                    & \multicolumn{2}{c}{Fairness ↓} \\ \cline{2-7} 
                      & Caucasian     & Asian         & African       & Indian        & STD           & AccGap        \\ \hline
        Best Threshold            & 0.6530        & 0.7300        & 0.5380        & 0.6660        & -             & -             \\ \hline
        Real Face                 & 0.7960/0.8964 & 0.7201/0.8813 & 0.8449/0.8767 & 0.7715/0.8888 & 0.0450/0.0075 & 0.1248/0.0197 \\
        FaceSwap                  & 0.8910/0.8265 & 0.8949/0.7883 & 0.7668/0.7373 & 0.8968/0.8264 & 0.0552/0.0366 & 0.1300/0.0892 \\ \hline
    \end{tabular}
\caption{Performance and fairness with 0.5 and optimal thresholds as threshold respectively. Left of '/' 0.5. Right of '/' is optimal threshold. }
\label{tab:threshold_sub1}
\end{table*}

\begin{table*}
    \centering
    \renewcommand{\arraystretch}{1.2}
    \tabcolsep=0.04cm
    \begin{tabular}{c|cccc|cccc}
        \hline
        Threshold & AADPD  & AADEOdds & AADEO  & AASTD  & URDPD  & URDEOdds & URDEO  & URSTD  \\ \hline
        0.5       & 0.1274 & 0.1274   & 0.1300 & 0.0501 & 0.1551 & 0.1551   & 0.1507 & 0.0607 \\ \hline
        BEST~\cite{robinson2020face}      & 0.0544 & 0.0544   & 0.0892 & 0.0220 & 0.0301 & 0.0301   & 0.0497 & 0.0122 \\ \hline
    \end{tabular}
\caption{Fairness metric results at different thresholds. }
\label{tab:threshold_sub2}
\end{table*}


Table~\ref{tab:balanced_training_dataset} presents the results. For convenience, we put the results of models trained on unbalanced and balanced datasets together. 
For both Naive Metric and Approach Averaged Metric, we observe that models trained on balanced datasets generally exhibit lower racial bias. Despite the significant improvement brought by a balanced dataset, these detectors still demonstrate a relatively high racial bias. It is crucial to note that, particularly for F3Net, not all metrics demonstrate improvement. These results indicate that training models on a balanced dataset cannot completely address the issue of racial bias, because there are other factors contributing to racial bias. 
For the Utility Regularized Metric, we note that the enhancement achieved through a balanced training dataset is less significant or may even diminish compared to the Naive Metric and Approach Averaged Metric. This is because this set of metrics is influenced by performance, and the reduced data volume results in a notable model performance drop. This highlights the advantage of the Utility Regularized Metric, i.e., it can reflect the model's performance. 

Based on the aforementioned observations, we argue that utilizing a race-balanced training dataset may not be a recommended way. The fact that deliberately collecting such balanced data in real-world scenarios usually implies discarding a wealth of available imbalanced datasets. Existing datasets are also seldom racially balanced. Furthermore, employing a racially balanced training set does not guarantee effective mitigation of racial bias. In conclusion, training with race-balanced datasets poses challenges, considering the scarcity of such datasets in real-world scenarios and the limited effectiveness in addressing racial bias. Therefore, it is preferable to utilize some other fair learning methods that exhibit better trade-offs.


\subsubsection{Setting Different Threshold for Each Race. }

The study in~\cite{robinson2020face} suggests that in face recognition, the confidence score distributions vary among different race groups. So, they propose setting different thresholds for different races can enhance fairness as well as improve overall performance. We utilize only the real face and FaceSwap portions of FairFD. Firstly, we plot the probability score distribution for each race in Figure~\ref{fig:probability_score}, revealing distinct differences in the confidence score distribution across different races. The Asian subset shows a higher frequency of high confidence scores, making it more prone to be classified as a fake face. Therefore, a larger threshold can be set for the Asian subset. On the other hand, the African subset exhibits a lower frequency of high probability scores, making it less likely to be classified as a fake face, allowing for a smaller threshold. 
Next, we set the optimal thresholds, which are values that maximize the overall performance for each racial subset. Table~\ref{tab:threshold_sub1} presents the optimal threshold values, as well as the performance and fairness scores when using threshold 0.5 and the optimal thresholds. We observe that, under the condition of maximizing overall performance, there is a certain degree of reduction in racial bias. We use test data as "Balanced Training Dataset". Table~\ref{tab:threshold_sub2} demonstrates racial bias using a fairness metric, showing that indeed, setting different thresholds can reduce racial bias. This conclusion aligns with the findings in~\cite{robinson2020face} consistently.

\subsubsection{Analysis of Frequency-Based Detector. } 

\begin{table*}[ht]
\centering
\renewcommand{\arraystretch}{1.1}  
\tabcolsep=0.05cm 
\begin{tabular}{cc|cc|cc|cc|cc}
\hline
\multicolumn{2}{c|}{\multirow{2}{*}{Fairness Metric}}                                              & \multicolumn{2}{c|}{Xception} & \multicolumn{2}{c|}{F3Net} & \multicolumn{2}{c|}{RECCE} & \multicolumn{2}{c}{UCF}  \\
\multicolumn{2}{c|}{}                                                                              & RGB       & Grayscale              & RGB              & Grayscale    & RGB     & Grayscale             & RGB    & Grayscale            \\ \hline
\multirow{4}{*}{\begin{tabular}[c]{@{}c@{}}Naive\\ Metric\end{tabular}}                 & DPD      & 0.1538    & \textbf{0.1309}   & \textbf{0.0764}  & 0.0780  & 0.1317  & \textbf{0.0811}  & 0.1782 & \textbf{0.1324} \\
                                                                                        & DEOdds   & 0.1672    & \textbf{0.1439}   & \textbf{0.0877}  & 0.0976  & 0.1379  & \textbf{0.0840}  & 0.1655 & \textbf{0.1512} \\
                                                                                        & DEO      & 0.2095    & \textbf{0.1789}   & \textbf{0.1030}  & 0.1035  & 0.1777  & \textbf{0.1076}  & 0.2334 & \textbf{0.1828} \\
                                                                                        & STD      & 0.0563    & \textbf{0.0470}   & \textbf{0.0278}  & 0.0283  & 0.0473  & \textbf{0.0295}  & 0.0635 & \textbf{0.0475} \\ \hline
\multirow{4}{*}{\begin{tabular}[c]{@{}c@{}}Approach\\ Averaged\\ Metric\end{tabular}}   & AADPD    & 0.1957    & \textbf{0.1673}   & \textbf{0.1102}  & 0.1147  & 0.1644  & \textbf{0.1097}  & 0.2107 & \textbf{0.1722} \\
                                                                                        & AADEOdds & 0.1674    & \textbf{0.1439}   & \textbf{0.0951}  & 0.1055  & 0.1378  & \textbf{0.0900}  & 0.1655 & \textbf{0.1512} \\
                                                                                        & AADEO    & 0.2099    & \textbf{0.1789}   & \textbf{0.1178}  & 0.1193  & 0.1777  & \textbf{0.1195}  & 0.2334 & \textbf{0.1828} \\
                                                                                        & AASTD    & 0.0721    & \textbf{0.0614}   & \textbf{0.0425}  & 0.0441  & 0.0603  & \textbf{0.0410}  & 0.0773 & \textbf{0.0658} \\ \hline
\multirow{4}{*}{\begin{tabular}[c]{@{}c@{}}Utility\\ Regularized\\ Metric\end{tabular}} & URDPD    & 0.1314    & \textbf{0.1144}   & \textbf{0.0769}  & 0.0802  & 0.1158  & \textbf{0.0785}  & 0.1433 & \textbf{0.1209} \\
                                                                                        & URDEOdds & 0.1069    & \textbf{0.0935}   & \textbf{0.0624}  & 0.0687  & 0.0908  & \textbf{0.0603}  & 0.1069 & \textbf{0.0986} \\
                                                                                        & URDEO    & 0.1437    & \textbf{0.1249}   & \textbf{0.0842}  & 0.0859  & 0.1283  & \textbf{0.0876}  & 0.1614 & \textbf{0.1321} \\
                                                                                        & URSTD    & 0.0482    & \textbf{0.0419}   & \textbf{0.0295}  & 0.0309  & 0.0423  & \textbf{0.0294}  & 0.0523 & \textbf{0.0461} \\ \hline
\end{tabular}
\caption{Comparison of fairness on RGB and Grayscale images. }
\label{tab:grayscale}
\end{table*}

In our previous findings, we observe that frequency-based methods exhibit lower racial bias compared to spatial-based methods. In this section, we provide a preliminary discussion using a subset of FairFD as "Balanced Training Dataset". Spatial-based methods focus on learning forgery clues in the spatial domain, mainly including color mismatch, textures, shapes, and blending boundaries~\cite{shiohara2022detecting}. On the other hand, frequency-based methods concentrate on learning forgery clues in the frequency domain, especially targeting high-frequency information related to blending boundaries, edges, and textures~\cite{qian2020thinking}. 
Frequency-based methods capture less color information, which is also crucial in distinguishing between different racial groups. Consequently, we make the assumption that frequency-based methods' racial biases are smaller due to these detectors learning less color information. 
Therefore, we convert the RGB images of our FairFD dataset into grayscale to eliminate color information but retain frequency domain information, and test multiple detectors. The results in Table~\ref{tab:grayscale} demonstrate that the racial bias of spatial-based detectors decreases, while the racial bias of frequency-based methods does not decrease and the changes are small. Thus, color indeed appears to be a contributing factor that leads spatial-based methods to have higher racial bias compared to frequency-based methods. 
There is still much exploration to be done in the comparative study of frequency and spatial domains, which can significantly contribute to the development of methods for mitigating racial bias. We leave this avenue of research for future investigation. 


\subsection{Visualization in Feature Space}

\begin{figure*}[t!]
\centering
    \includegraphics[width=0.967\linewidth]{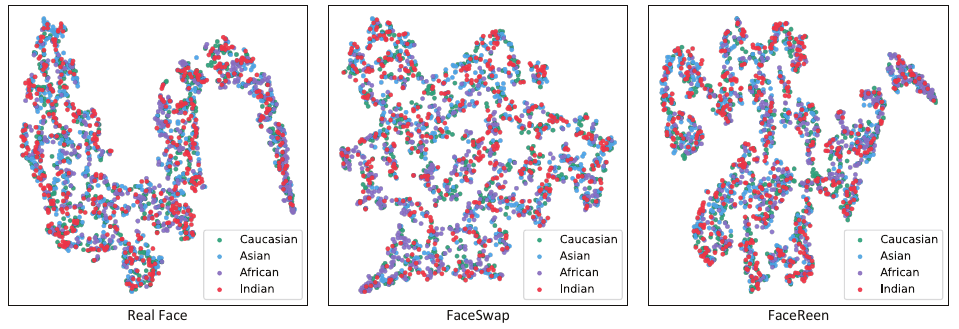}
\caption{T-SNE visualization on Xception model. }
\label{fig:vis_feature}
\end{figure*}
\begin{figure*}[t!]
\centering
    \includegraphics[width=0.967\linewidth]{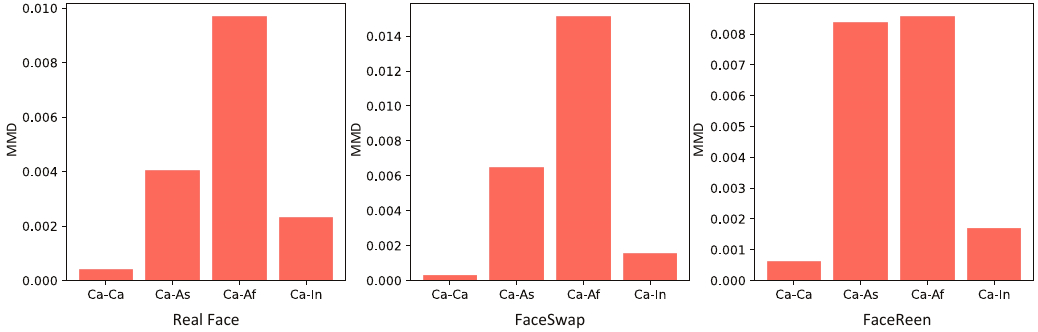}
\caption{Maximum Mean Discrepancy (MMD) on Xception model. }
\label{fig:mmd_feature}
\end{figure*}

The research on fairness in face forgery detection is similar to the study of fairness in face recognition(see section "Fairness in Face Recognition" in \textit{Supp} for more details). As stated in~\cite{wang2019racial}, one reason for racial bias is that different subsets' features are totally separate. So they take the racial bias as a problem of domain gap and propose IMAN(information maximization adaptation network) to decrease this domain gap. 
We sample 500 samples for each ethnic group and use the well trained Xception model as the detector to obtain features for these samples. Then, we plot the t-SNE dimensionality reduction graphs for the feature spaces of the four ethnic subsets in Figure~\ref{fig:vis_feature}. Considering that different forgery approaches will cause distinct results, we plot for Real Face, FaceSwap and FaceReen respectively. Unlike in~\cite{wang2019racial}, we do not observe distinct separation between different subsets at feature level. 
Next, we use the MMD(Maximum Mean Discrepancy) to mathematically calculate the feature distances between the Caucasian subset and other ethnic groups. The results in Figure~\ref{fig:mmd_feature} demonstrate that although distances show difference, in comparison to the distances in~\cite{wang2019racial}, the distances here are all nearly close to zero. 
These results are because they consider the face recognition task in~\cite{wang2019racial}, so the models tends to learn distinctive features for each subject, resulting in significant differences at the feature level. In contrast, for the task of face forgery detection, the detector does not exhibit the same tendency, leading to smaller differences in feature level. Therefore, attempting to enhance the fairness of face forgery detection from a feature-level perspective may not be feasible.

\subsection{Results without Data Augmentation}

Results without data augmentation are shown in Table~\ref{tab:fairness-enhanced-no-aug}. 

\begin{table}[t!]
\centering
\renewcommand{\arraystretch}{1.0}  
\tabcolsep=0.1cm 
\begin{tabular}{cccc}
\toprule
\multicolumn{1}{c}{\multirow{2}{*}{Fairness Metric}} & \multicolumn{3}{c}{Fairness-enhanced} \\ \cmidrule(r){2-4}
\multicolumn{1}{c}{} & DAG & DAW & PFGDFD \\ \midrule \midrule
DPD$\downarrow$      & 0.0316 & 0.0694 & 0.0709 \\
DEOdds$\downarrow$   & 0.0564 & 0.1057 & 0.1102 \\
DEO$\downarrow$      & 0.0402 & 0.0870 & 0.0893 \\
STD$\downarrow$      & 0.0122 & 0.0253 & 0.0280 \\ \midrule
AADPD$\downarrow$    & 0.0569 & 0.0989 & 0.1127 \\
AADEOdds$\downarrow$ & 0.0641 & 0.1105 & 0.1210 \\
AADEO$\downarrow$    & 0.0555 & 0.0966 & 0.1110 \\
AASTD$\downarrow$    & 0.0225 & 0.0378 & 0.0436 \\ \midrule
URDPD$\downarrow$    & 0.0322 & 0.0574 & 0.0688 \\
URDEOdds$\downarrow$ & 0.0447 & 0.0767 & 0.0763 \\
URDEO$\downarrow$    & 0.0297 & 0.0536 & 0.0673 \\ 
URSTD$\downarrow$    & 0.0127 & 0.0219 & 0.0266 \\ \midrule
AUC$\uparrow$        & 0.6638 & 0.6121 & 0.6336 \\ \bottomrule
\end{tabular}
\caption{Bias evaluation for three detectors trained without data augmentation.}
\label{tab:fairness-enhanced-no-aug}
\end{table}


\begin{table*}[ht]
\centering
\renewcommand{\arraystretch}{1.3}  
\tabcolsep=0.018cm 
\resizebox{1.0\linewidth}{!}{
\begin{tabular}{c|c|cccc|cccc|cccc|cc}
\toprule
                               &                              & \multicolumn{4}{c|}{Naive Metric$\downarrow$}                                                                                                  & \multicolumn{4}{c|}{Approach Averaged Metric$\downarrow$}                                                                                        & \multicolumn{4}{c|}{Utility Regularized Metric$\downarrow$}                                                                                & \multicolumn{2}{c}{Utility$\uparrow$}                                     \\ \cline{3-16} 
\multirow{-2}{*}{\begin{tabular}[c]{@{}c@{}}Pruning\\ Rate\end{tabular}} & \multirow{-2}{*}{Method}     & DPD                                     & DEOdds                                  & DEO                                     & STD                                     & \begin{tabular}[c]{@{}c@{}}MA\\ DPD\end{tabular}                                   & \begin{tabular}[c]{@{}c@{}}MA\\ DEOdds\end{tabular}                                & \begin{tabular}[c]{@{}c@{}}MA\\ DEO\end{tabular}                                   & \begin{tabular}[c]{@{}c@{}}MA\\ STD\end{tabular}                                   & \begin{tabular}[c]{@{}c@{}}UR\\ DPD\end{tabular}                                   & \begin{tabular}[c]{@{}c@{}}UR\\ DEOdds\end{tabular}                                & \begin{tabular}[c]{@{}c@{}}UR\\ DEO\end{tabular}                                   & \begin{tabular}[c]{@{}c@{}}UR\\ STD\end{tabular}                                   & AUC                                     & ACC                                     \\ \hline
0                              & Original                     & 0.0203                         & 0.0304                         & 0.0215                         & 0.0080                         & 0.0556                         & 0.0481                         & 0.0571                         & 0.0219                         & 0.0320                         & 0.0299                         & 0.0324                         & 0.0126                         & 0.6763                         & 0.7618                         \\ \hline
                               & WEIG                         & 0.0183                         & 0.0258                         & 0.0201                         & 0.0072                         & 0.0564                         & 0.0451                         & 0.0586                         & 0.0219                         & 0.0324                         & 0.0277                         & 0.0334                         & 0.0126                         & 0.6769                         & 0.7615                         \\
                               & RoBA                         & 0.0188                         & 0.0258                         & 0.0214                         & 0.0069                         & 0.0602                         & 0.0466                         & 0.0629                         & 0.0233                         & 0.0458                         & 0.0322                         & 0.0486                         & 0.0177                         & 0.6608                         & 0.3468                         \\
\multirow{-3}{*}{0.1\%}        & \cellcolor[HTML]{E7E6E6}BPFA & \cellcolor[HTML]{E7E6E6}0.0183 & \cellcolor[HTML]{E7E6E6}0.0258 & \cellcolor[HTML]{E7E6E6}0.0201 & \cellcolor[HTML]{E7E6E6}0.0072 & \cellcolor[HTML]{E7E6E6}0.0564 & \cellcolor[HTML]{E7E6E6}0.0451 & \cellcolor[HTML]{E7E6E6}0.0586 & \cellcolor[HTML]{E7E6E6}0.0219 & \cellcolor[HTML]{E7E6E6}0.0324 & \cellcolor[HTML]{E7E6E6}0.0277 & \cellcolor[HTML]{E7E6E6}0.0334 & \cellcolor[HTML]{E7E6E6}0.0126 & \cellcolor[HTML]{E7E6E6}0.6769 & \cellcolor[HTML]{E7E6E6}0.7615 \\ \midrule
                               & WEIG                         & 0.0184                         & 0.0259                         & 0.0201                         & 0.0072                         & 0.0563                         & 0.0451                         & 0.0586                         & 0.0219                         & 0.0324                         & 0.0277                         & 0.0334                         & 0.0126                         & 0.6769                         & 0.7615                         \\
                               & RoBA                         & 0.1128                         & 0.1598                         & 0.1395                         & 0.0445                         & 0.1462                         & 0.1616                         & 0.1432                         & 0.0583                         & 0.0893                         & 0.1024                         & 0.0867                         & 0.0356                         & 0.6331                         & 0.7037                         \\
\multirow{-3}{*}{0.4\%}        & \cellcolor[HTML]{E7E6E6}BPFA & \cellcolor[HTML]{E7E6E6}0.0184 & \cellcolor[HTML]{E7E6E6}0.0259 & \cellcolor[HTML]{E7E6E6}0.0201 & \cellcolor[HTML]{E7E6E6}0.0072 & \cellcolor[HTML]{E7E6E6}0.0563 & \cellcolor[HTML]{E7E6E6}0.0451 & \cellcolor[HTML]{E7E6E6}0.0586 & \cellcolor[HTML]{E7E6E6}0.0219 & \cellcolor[HTML]{E7E6E6}0.0324 & \cellcolor[HTML]{E7E6E6}0.0277 & \cellcolor[HTML]{E7E6E6}0.0334 & \cellcolor[HTML]{E7E6E6}0.0126 & \cellcolor[HTML]{E7E6E6}0.6770 & \cellcolor[HTML]{E7E6E6}0.7613 \\ \midrule
                               & WEIG                         & 0.0181                         & 0.0259                         & 0.0199                         & 0.0071                         & 0.0563                         & 0.0452                         & 0.0585                         & 0.0219                         & 0.0324                         & 0.0278                         & 0.0333                         & 0.0126                         & 0.6769                         & 0.7613                         \\
                               & RoBA                         & 0.0210                         & 0.0166                         & 0.0234                         & 0.0077                         & 0.0268                         & 0.0191                         & 0.0283                         & 0.0098                         & 0.0224                         & 0.0145                         & 0.0240                         & 0.0082                         & 0.6853                         & 0.1760                         \\
\multirow{-3}{*}{0.7\%}        & \cellcolor[HTML]{E7E6E6}BPFA & \cellcolor[HTML]{E7E6E6}0.0181 & \cellcolor[HTML]{E7E6E6}0.0259 & \cellcolor[HTML]{E7E6E6}0.0199 & \cellcolor[HTML]{E7E6E6}0.0071 & \cellcolor[HTML]{E7E6E6}0.0563 & \cellcolor[HTML]{E7E6E6}0.0452 & \cellcolor[HTML]{E7E6E6}0.0585 & \cellcolor[HTML]{E7E6E6}0.0219 & \cellcolor[HTML]{E7E6E6}0.0324 & \cellcolor[HTML]{E7E6E6}0.0278 & \cellcolor[HTML]{E7E6E6}0.0333 & \cellcolor[HTML]{E7E6E6}0.0126 & \cellcolor[HTML]{E7E6E6}0.6769 & \cellcolor[HTML]{E7E6E6}0.7615 \\ \midrule
                               & WEIG                         & 0.0189                         & 0.0296                         & 0.0201                         & 0.0075                         & 0.0562                         & 0.0484                         & 0.0577                         & 0.0220                         & 0.0323                         & 0.0300                         & 0.0328                         & 0.0126                         & 0.6760                         & 0.7603                         \\
                               & RoBA                         & 0.0605                         & 0.0496                         & 0.0686                         & 0.0239                         & 0.0777                         & 0.0563                         & 0.0819                         & 0.0300                         & 0.0616                         & 0.0408                         & 0.0658                         & 0.0238                         & 0.6913                         & 0.3124                         \\
\multirow{-3}{*}{1\%}          & \cellcolor[HTML]{E7E6E6}BPFA & \cellcolor[HTML]{E7E6E6}0.0195 & \cellcolor[HTML]{E7E6E6}0.0295 & \cellcolor[HTML]{E7E6E6}0.0208 & \cellcolor[HTML]{E7E6E6}0.0079 & \cellcolor[HTML]{E7E6E6}0.0556 & \cellcolor[HTML]{E7E6E6}0.0477 & \cellcolor[HTML]{E7E6E6}0.0571 & \cellcolor[HTML]{E7E6E6}0.0218 & \cellcolor[HTML]{E7E6E6}0.0320 & \cellcolor[HTML]{E7E6E6}0.0296 & \cellcolor[HTML]{E7E6E6}0.0325 & \cellcolor[HTML]{E7E6E6}0.0125 & \cellcolor[HTML]{E7E6E6}0.6766 & \cellcolor[HTML]{E7E6E6}0.7611 \\ \midrule
                               & WEIG                         & 0.0190                         & 0.0277                         & 0.0211                         & 0.0074                         & 0.0569                         & 0.0467                         & 0.0590                         & 0.0221                         & 0.0328                         & 0.0288                         & 0.0336                         & 0.0128                         & 0.6766                         & 0.7596                         \\
                               & RoBA                         & -                              & -                              & -                              & -                              & -                              & -                              & -                              & -                              & -                              & -                              & -                              & -                              & -                              & -                              \\
\multirow{-3}{*}{4\%}          & \cellcolor[HTML]{E7E6E6}BPFA & \cellcolor[HTML]{E7E6E6}0.0164 & \cellcolor[HTML]{E7E6E6}0.0242 & \cellcolor[HTML]{E7E6E6}0.0179 & \cellcolor[HTML]{E7E6E6}0.0063 & \cellcolor[HTML]{E7E6E6}0.0523 & \cellcolor[HTML]{E7E6E6}0.0424 & \cellcolor[HTML]{E7E6E6}0.0543 & \cellcolor[HTML]{E7E6E6}0.0203 & \cellcolor[HTML]{E7E6E6}0.0298 & \cellcolor[HTML]{E7E6E6}0.0262 & \cellcolor[HTML]{E7E6E6}0.0305 & \cellcolor[HTML]{E7E6E6}0.0116 & \cellcolor[HTML]{E7E6E6}0.6788 & \cellcolor[HTML]{E7E6E6}0.7813 \\ \midrule
                               & WEIG                         & 0.0196                         & 0.0274                         & 0.0217                         & 0.0076                         & 0.0564                         & 0.0459                         & 0.0586                         & 0.0220                         & 0.0326                         & 0.0282                         & 0.0334                         & 0.0127                         & 0.6777                         & 0.7587                         \\
                               & RoBA                         & -                              & -                              & -                              & -                              & -                              & -                              & -                              & -                              & -                              & -                              & -                              & -                              & -                              & -                              \\
\multirow{-3}{*}{7\%}          & \cellcolor[HTML]{E7E6E6}BPFA & \cellcolor[HTML]{E7E6E6}0.0181 & \cellcolor[HTML]{E7E6E6}0.0209 & \cellcolor[HTML]{E7E6E6}0.0200 & \cellcolor[HTML]{E7E6E6}0.0072 & \cellcolor[HTML]{E7E6E6}0.0473 & \cellcolor[HTML]{E7E6E6}0.0357 & \cellcolor[HTML]{E7E6E6}0.0496 & \cellcolor[HTML]{E7E6E6}0.0182 & \cellcolor[HTML]{E7E6E6}0.0265 & \cellcolor[HTML]{E7E6E6}0.0218 & \cellcolor[HTML]{E7E6E6}0.0275 & \cellcolor[HTML]{E7E6E6}0.0102 & \cellcolor[HTML]{E7E6E6}0.6862 & \cellcolor[HTML]{E7E6E6}0.8055 \\ \midrule
                               & WEIG                         & 0.0195                         & 0.0268                         & 0.0215                         & 0.0076                         & 0.0560                         & 0.0451                         & 0.0582                         & 0.0219                         & 0.0323                         & 0.0277                         & 0.0332                         & 0.0126                         & 0.6777                         & 0.7599                         \\
                               & RoBA                         & -                              & -                              & -                              & -                              & -                              & -                              & -                              & -                              & -                              & -                              & -                              & -                              & -                              & -                              \\
\multirow{-3}{*}{10\%}         & \cellcolor[HTML]{E7E6E6}BPFA & \cellcolor[HTML]{E7E6E6}0.0239 & \cellcolor[HTML]{E7E6E6}0.0178 & \cellcolor[HTML]{E7E6E6}0.0259 & \cellcolor[HTML]{E7E6E6}0.0096 & \cellcolor[HTML]{E7E6E6}0.0487 & \cellcolor[HTML]{E7E6E6}0.0310 & \cellcolor[HTML]{E7E6E6}0.0523 & \cellcolor[HTML]{E7E6E6}0.0196 & \cellcolor[HTML]{E7E6E6}0.0274 & \cellcolor[HTML]{E7E6E6}0.0181 & \cellcolor[HTML]{E7E6E6}0.0293 & \cellcolor[HTML]{E7E6E6}0.0110 & \cellcolor[HTML]{E7E6E6}0.6938 & \cellcolor[HTML]{E7E6E6}0.7900 \\ \bottomrule
\end{tabular}}
\caption{Ablation study on pruning rate of SPSL. We use '-' to indicate that the model is completely unusable (AUC = 0.5). }
\label{tab:pruning_rate_spsl}
\end{table*}

\begin{table*}[ht]
\centering
\renewcommand{\arraystretch}{1.3}  
\tabcolsep=0.018cm 
\resizebox{1.0\linewidth}{!}{
\begin{tabular}{c|c|cccc|cccc|cccc|cc}
\toprule
                               &                              & \multicolumn{4}{c|}{Naive Metric}                                                                                                  & \multicolumn{4}{c|}{Approach Averaged Metric}                                                                                        & \multicolumn{4}{c|}{Utility Regularized Metric}                                                                                & \multicolumn{2}{c}{Utility}                                     \\ \cline{3-16} 
\multirow{-2}{*}{\begin{tabular}[c]{@{}c@{}}Pruning\\ Rate\end{tabular}} & \multirow{-2}{*}{Method}     & DPD                                     & DEOdds                                  & DEO                                     & STD                                     & \begin{tabular}[c]{@{}c@{}}MA\\ DPD\end{tabular}                                   & \begin{tabular}[c]{@{}c@{}}MA\\ DEOdds\end{tabular}                                & \begin{tabular}[c]{@{}c@{}}MA\\ DEO\end{tabular}                                   & \begin{tabular}[c]{@{}c@{}}MA\\ STD\end{tabular}                                   & \begin{tabular}[c]{@{}c@{}}UR\\ DPD\end{tabular}                                   & \begin{tabular}[c]{@{}c@{}}UR\\ DEOdds\end{tabular}                                & \begin{tabular}[c]{@{}c@{}}UR\\ DEO\end{tabular}                                   & \begin{tabular}[c]{@{}c@{}}UR\\ STD\end{tabular}                                   & AUC                                     & ACC                                     \\ \hline
0                              & Original                     & 0.1099                         & 0.1005                         & 0.1242                         & 0.0398                         & 0.1552                         & 0.1196                         & 0.1623                         & 0.0576                         & 0.1037                         & 0.0763                         & 0.1092                         & 0.0384                         & 0.7304                         & 0.5751                         \\ \hline
                               & WEIG                         & 0.1099                         & 0.1006                         & 0.1241                         & 0.0398                         & 0.1552                         & 0.1196                         & 0.1623                         & 0.0576                         & 0.1037                         & 0.0763                         & 0.1091                         & 0.0384                         & 0.7304                         & 0.5750                         \\
                               & RoBA                         & 0.0369                         & 0.0343                         & 0.0415                         & 0.0151                         & 0.0676                         & 0.0492                         & 0.0713                         & 0.0265                         & 0.0566                         & 0.0371                         & 0.0604                         & 0.0221                         & 0.5967                         & 0.2235                         \\
\multirow{-3}{*}{0.1\%}        & \cellcolor[HTML]{E7E6E6}BPFA & \cellcolor[HTML]{E7E6E6}0.1099 & \cellcolor[HTML]{E7E6E6}0.1006 & \cellcolor[HTML]{E7E6E6}0.1241 & \cellcolor[HTML]{E7E6E6}0.0398 & \cellcolor[HTML]{E7E6E6}0.1552 & \cellcolor[HTML]{E7E6E6}0.1197 & \cellcolor[HTML]{E7E6E6}0.1623 & \cellcolor[HTML]{E7E6E6}0.0576 & \cellcolor[HTML]{E7E6E6}0.1037 & \cellcolor[HTML]{E7E6E6}0.0763 & \cellcolor[HTML]{E7E6E6}0.1092 & \cellcolor[HTML]{E7E6E6}0.0384 & \cellcolor[HTML]{E7E6E6}0.7305 & \cellcolor[HTML]{E7E6E6}0.5751 \\ \midrule
                               & WEIG                         & 0.1099                         & 0.1006                         & 0.1242                         & 0.0398                         & 0.1552                         & 0.1196                         & 0.1623                         & 0.0576                         & 0.1037                         & 0.0763                         & 0.1091                         & 0.0384                         & 0.7304                         & 0.5751                         \\
                               & RoBA                         & 0.0582                         & 0.0649                         & 0.0690                         & 0.0223                         & 0.0763                         & 0.0693                         & 0.0777                         & 0.0299                         & 0.0607                         & 0.0475                         & 0.0634                         & 0.0238                         & 0.5706                         & 0.2187                         \\
\multirow{-3}{*}{0.4\%}        & \cellcolor[HTML]{E7E6E6}BPFA & \cellcolor[HTML]{E7E6E6}0.1100 & \cellcolor[HTML]{E7E6E6}0.1003 & \cellcolor[HTML]{E7E6E6}0.1242 & \cellcolor[HTML]{E7E6E6}0.0398 & \cellcolor[HTML]{E7E6E6}0.1552 & \cellcolor[HTML]{E7E6E6}0.1194 & \cellcolor[HTML]{E7E6E6}0.1623 & \cellcolor[HTML]{E7E6E6}0.0576 & \cellcolor[HTML]{E7E6E6}0.1037 & \cellcolor[HTML]{E7E6E6}0.0762 & \cellcolor[HTML]{E7E6E6}0.1092 & \cellcolor[HTML]{E7E6E6}0.0384 & \cellcolor[HTML]{E7E6E6}0.7305 & \cellcolor[HTML]{E7E6E6}0.5751 \\ \midrule
                               & WEIG                         & 0.1100                         & 0.1005                         & 0.1242                         & 0.0398                         & 0.1553                         & 0.1196                         & 0.1624                         & 0.0577                         & 0.1037                         & 0.0763                         & 0.1092                         & 0.0384                         & 0.7304                         & 0.5750                         \\
                               & RoBA                         & 0.0467                         & 0.0600                         & 0.0567                         & 0.0191                         & 0.0654                         & 0.0645                         & 0.0656                         & 0.0261                         & 0.0390                         & 0.0445                         & 0.0379                         & 0.0156                         & 0.5690                         & 0.7860                         \\
\multirow{-3}{*}{0.7\%}        & \cellcolor[HTML]{E7E6E6}BPFA & \cellcolor[HTML]{E7E6E6}0.1098 & \cellcolor[HTML]{E7E6E6}0.1000 & \cellcolor[HTML]{E7E6E6}0.1239 & \cellcolor[HTML]{E7E6E6}0.0398 & \cellcolor[HTML]{E7E6E6}0.1547 & \cellcolor[HTML]{E7E6E6}0.1190 & \cellcolor[HTML]{E7E6E6}0.1619 & \cellcolor[HTML]{E7E6E6}0.0575 & \cellcolor[HTML]{E7E6E6}0.1033 & \cellcolor[HTML]{E7E6E6}0.0759 & \cellcolor[HTML]{E7E6E6}0.1088 & \cellcolor[HTML]{E7E6E6}0.0383 & \cellcolor[HTML]{E7E6E6}0.7303 & \cellcolor[HTML]{E7E6E6}0.5757 \\ \midrule
                               & WEIG                         & 0.1098                         & 0.1003                         & 0.1240                         & 0.0398                         & 0.1550                         & 0.1194                         & 0.1621                         & 0.0576                         & 0.1035                         & 0.0761                         & 0.1090                         & 0.0384                         & 0.7304                         & 0.5751                         \\
                               & RoBA                         & 0.0400                         & 0.0621                         & 0.0503                         & 0.0155                         & 0.0556                         & 0.0639                         & 0.0539                         & 0.0219                         & 0.0327                         & 0.0465                         & 0.0299                         & 0.0128                         & 0.5670                         & 0.8283                         \\
\multirow{-3}{*}{1\%}          & \cellcolor[HTML]{E7E6E6}BPFA & \cellcolor[HTML]{E7E6E6}0.1096 & \cellcolor[HTML]{E7E6E6}0.0999 & \cellcolor[HTML]{E7E6E6}0.1237 & \cellcolor[HTML]{E7E6E6}0.0397 & \cellcolor[HTML]{E7E6E6}0.1546 & \cellcolor[HTML]{E7E6E6}0.1189 & \cellcolor[HTML]{E7E6E6}0.1617 & \cellcolor[HTML]{E7E6E6}0.0574 & \cellcolor[HTML]{E7E6E6}0.1032 & \cellcolor[HTML]{E7E6E6}0.0758 & \cellcolor[HTML]{E7E6E6}0.1087 & \cellcolor[HTML]{E7E6E6}0.0382 & \cellcolor[HTML]{E7E6E6}0.7305 & \cellcolor[HTML]{E7E6E6}0.5760 \\ \midrule
                               & WEIG                         & 0.1105                         & 0.1008                         & 0.1249                         & 0.0400                         & 0.1560                         & 0.1199                         & 0.1632                         & 0.0579                         & 0.1042                         & 0.0765                         & 0.1097                         & 0.0386                         & 0.7307                         & 0.5752                         \\
                               & RoBA                         & -                              & -                              & -                              & -                              & -                              & -                              & -                              & -                              & -                              & -                              & -                              & -                              & -                              & -                              \\
\multirow{-3}{*}{4\%}          & \cellcolor[HTML]{E7E6E6}BPFA & \cellcolor[HTML]{E7E6E6}0.1099 & \cellcolor[HTML]{E7E6E6}0.1016 & \cellcolor[HTML]{E7E6E6}0.1244 & \cellcolor[HTML]{E7E6E6}0.0397 & \cellcolor[HTML]{E7E6E6}0.1551 & \cellcolor[HTML]{E7E6E6}0.1204 & \cellcolor[HTML]{E7E6E6}0.1621 & \cellcolor[HTML]{E7E6E6}0.0575 & \cellcolor[HTML]{E7E6E6}0.1032 & \cellcolor[HTML]{E7E6E6}0.0765 & \cellcolor[HTML]{E7E6E6}0.1085 & \cellcolor[HTML]{E7E6E6}0.0382 & \cellcolor[HTML]{E7E6E6}0.7314 & \cellcolor[HTML]{E7E6E6}0.5812 \\ \midrule
                               & WEIG                         & 0.1110                         & 0.1026                         & 0.1260                         & 0.0402                         & 0.1578                         & 0.1221                         & 0.1649                         & 0.0585                         & 0.1061                         & 0.0781                         & 0.1117                         & 0.0392                         & 0.7317                         & 0.5678                         \\
                               & RoBA                         & -                              & -                              & -                              & -                              & -                              & -                              & -                              & -                              & -                              & -                              & -                              & -                              & -                              & -                              \\
\multirow{-3}{*}{7\%}          & \cellcolor[HTML]{E7E6E6}BPFA & \cellcolor[HTML]{E7E6E6}0.1029 & \cellcolor[HTML]{E7E6E6}0.1153 & \cellcolor[HTML]{E7E6E6}0.1168 & \cellcolor[HTML]{E7E6E6}0.0372 & \cellcolor[HTML]{E7E6E6}0.1562 & \cellcolor[HTML]{E7E6E6}0.1370 & \cellcolor[HTML]{E7E6E6}0.1600 & \cellcolor[HTML]{E7E6E6}0.0582 & \cellcolor[HTML]{E7E6E6}0.1064 & \cellcolor[HTML]{E7E6E6}0.0871 & \cellcolor[HTML]{E7E6E6}0.1103 & \cellcolor[HTML]{E7E6E6}0.0395 & \cellcolor[HTML]{E7E6E6}0.7151 & \cellcolor[HTML]{E7E6E6}0.5413 \\ \midrule
                               & WEIG                         & 0.1114                         & 0.1012                         & 0.1264                         & 0.0402                         & 0.1579                         & 0.1207                         & 0.1654                         & 0.0586                         & 0.1070                         & 0.0777                         & 0.1129                         & 0.0396                         & 0.7300                         & 0.5574                         \\
                               & RoBA                         & -                              & -                              & -                              & -                              & -                              & -                              & -                              & -                              & -                              & -                              & -                              & -                              & -                              & -                              \\
\multirow{-3}{*}{10\%}         & \cellcolor[HTML]{E7E6E6}BPFA & \cellcolor[HTML]{E7E6E6}0.0946 & \cellcolor[HTML]{E7E6E6}0.0822 & \cellcolor[HTML]{E7E6E6}0.1071 & \cellcolor[HTML]{E7E6E6}0.0351 & \cellcolor[HTML]{E7E6E6}0.1453 & \cellcolor[HTML]{E7E6E6}0.1053 & \cellcolor[HTML]{E7E6E6}0.1533 & \cellcolor[HTML]{E7E6E6}0.0542 & \cellcolor[HTML]{E7E6E6}0.1048 & \cellcolor[HTML]{E7E6E6}0.0711 & \cellcolor[HTML]{E7E6E6}0.1115 & \cellcolor[HTML]{E7E6E6}0.0389 & \cellcolor[HTML]{E7E6E6}0.7312 & \cellcolor[HTML]{E7E6E6}0.4728 \\ \bottomrule
\end{tabular}}
\caption{Ablation study on pruning rate of FFD. We use '-' to indicate that the model is completely unusable (AUC = 0.5). }
\label{tab:pruning_rate_ffd}
\end{table*}

\begin{table*}[ht]
\centering
\renewcommand{\arraystretch}{1.3}  
\tabcolsep=0.018cm 
\resizebox{1.0\linewidth}{!}{
\begin{tabular}{c|c|cccc|cccc|cccc|cc}
\toprule
                               &                              & \multicolumn{4}{c|}{Naive Metric$\downarrow$}                                                                                                  & \multicolumn{4}{c|}{Approach Averaged Metric$\downarrow$}                                                                                        & \multicolumn{4}{c|}{Utility Regularized Metric$\downarrow$}                                                                                & \multicolumn{2}{c}{Utility$\uparrow$}                                     \\ \cline{3-16} 
\multirow{-2}{*}{\begin{tabular}[c]{@{}c@{}}Pruning\\ Rate\end{tabular}} & \multirow{-2}{*}{Method}     & DPD                                     & DEOdds                                  & DEO                                     & STD                                     & \begin{tabular}[c]{@{}c@{}}MA\\ DPD\end{tabular}                                   & \begin{tabular}[c]{@{}c@{}}MA\\ DEOdds\end{tabular}                                & \begin{tabular}[c]{@{}c@{}}MA\\ DEO\end{tabular}                                   & \begin{tabular}[c]{@{}c@{}}MA\\ STD\end{tabular}                                   & \begin{tabular}[c]{@{}c@{}}UR\\ DPD\end{tabular}                                   & \begin{tabular}[c]{@{}c@{}}UR\\ DEOdds\end{tabular}                                & \begin{tabular}[c]{@{}c@{}}UR\\ DEO\end{tabular}                                   & \begin{tabular}[c]{@{}c@{}}UR\\ STD\end{tabular}                                   & AUC                                     & ACC                                     \\ \hline
0                              & Original                     & 0.0805                         & 0.1396                         & 0.1032                         & 0.0328                         & 0.1393                         & 0.1560                         & 0.1360                         & 0.0530                         & 0.0881                         & 0.0986                         & 0.0860                         & 0.0335                         & 0.6302                         & 0.6019                         \\ \hline
                               & WEIG                         & 0.0789                         & 0.1340                         & 0.1013                         & 0.0319                         & 0.1349                         & 0.1494                         & 0.1320                         & 0.0513                         & 0.0853                         & 0.0945                         & 0.0835                         & 0.0324                         & 0.6298                         & 0.6021                         \\
                               & RoBA                         & 0.0216                         & 0.0369                         & 0.0278                         & 0.0078                         & 0.0316                         & 0.0381                         & 0.0303                         & 0.0122                         & 0.0181                         & 0.0296                         & 0.0158                         & 0.0070                         & 0.5468                         & 0.8798                         \\
\multirow{-3}{*}{0.1\%}        & \cellcolor[HTML]{E7E6E6}BPFA & \cellcolor[HTML]{E7E6E6}0.0785 & \cellcolor[HTML]{E7E6E6}0.1392 & \cellcolor[HTML]{E7E6E6}0.1010 & \cellcolor[HTML]{E7E6E6}0.0320 & \cellcolor[HTML]{E7E6E6}0.1366 & \cellcolor[HTML]{E7E6E6}0.1552 & \cellcolor[HTML]{E7E6E6}0.1329 & \cellcolor[HTML]{E7E6E6}0.0521 & \cellcolor[HTML]{E7E6E6}0.0864 & \cellcolor[HTML]{E7E6E6}0.0982 & \cellcolor[HTML]{E7E6E6}0.0840 & \cellcolor[HTML]{E7E6E6}0.0329 & \cellcolor[HTML]{E7E6E6}0.6300 & \cellcolor[HTML]{E7E6E6}0.6039 \\ \hline
                               & WEIG                         & 0.0789                         & 0.1340                         & 0.1012                         & 0.0319                         & 0.1349                         & 0.1494                         & 0.1320                         & 0.0513                         & 0.0853                         & 0.0944                         & 0.0835                         & 0.0324                         & 0.6298                         & 0.6021                         \\
                               & RoBA                         & -                              & -                              & -                              & -                              & -                              & -                              & -                              & -                              & -                              & -                              & -                              & -                              & -                              & -                              \\
\multirow{-3}{*}{0.4\%}        & \cellcolor[HTML]{E7E6E6}BPFA & \cellcolor[HTML]{E7E6E6}0.0594 & \cellcolor[HTML]{E7E6E6}0.1337 & \cellcolor[HTML]{E7E6E6}0.0796 & \cellcolor[HTML]{E7E6E6}0.0238 & \cellcolor[HTML]{E7E6E6}0.1079 & \cellcolor[HTML]{E7E6E6}0.1442 & \cellcolor[HTML]{E7E6E6}0.1006 & \cellcolor[HTML]{E7E6E6}0.0411 & \cellcolor[HTML]{E7E6E6}0.0644 & \cellcolor[HTML]{E7E6E6}0.0969 & \cellcolor[HTML]{E7E6E6}0.0578 & \cellcolor[HTML]{E7E6E6}0.0245 & \cellcolor[HTML]{E7E6E6}0.6445 & \cellcolor[HTML]{E7E6E6}0.7415 \\ \hline
                               & WEIG                         & 0.0789                         & 0.1340                         & 0.1012                         & 0.0319                         & 0.1349                         & 0.1494                         & 0.1320                         & 0.0513                         & 0.0853                         & 0.0945                         & 0.0835                         & 0.0324                         & 0.6298                         & 0.6021                         \\
                               & RoBA                         & -                              & -                              & -                              & -                              & -                              & -                              & -                              & -                              & -                              & -                              & -                              & -                              & -                              & -                              \\
\multirow{-3}{*}{0.7\%}        & \cellcolor[HTML]{E7E6E6}BPFA & \cellcolor[HTML]{E7E6E6}-      & \cellcolor[HTML]{E7E6E6}-      & \cellcolor[HTML]{E7E6E6}-      & \cellcolor[HTML]{E7E6E6}-      & \cellcolor[HTML]{E7E6E6}-      & \cellcolor[HTML]{E7E6E6}-      & \cellcolor[HTML]{E7E6E6}-      & \cellcolor[HTML]{E7E6E6}-      & \cellcolor[HTML]{E7E6E6}-      & \cellcolor[HTML]{E7E6E6}-      & \cellcolor[HTML]{E7E6E6}-      & \cellcolor[HTML]{E7E6E6}-      & \cellcolor[HTML]{E7E6E6}-      & \cellcolor[HTML]{E7E6E6}-      \\ \hline
                               & WEIG                         & 0.0789                         & 0.1340                         & 0.1012                         & 0.0319                         & 0.1349                         & 0.1494                         & 0.1320                         & 0.0513                         & 0.0853                         & 0.0945                         & 0.0835                         & 0.0324                         & 0.6298                         & 0.6021                         \\
                               & RoBA                         & -                              & -                              & -                              & -                              & -                              & -                              & -                              & -                              & -                              & -                              & -                              & -                              & -                              & -                              \\
\multirow{-3}{*}{1\%}          & \cellcolor[HTML]{E7E6E6}BPFA & \cellcolor[HTML]{E7E6E6}-      & \cellcolor[HTML]{E7E6E6}-      & \cellcolor[HTML]{E7E6E6}-      & \cellcolor[HTML]{E7E6E6}-      & \cellcolor[HTML]{E7E6E6}-      & \cellcolor[HTML]{E7E6E6}-      & \cellcolor[HTML]{E7E6E6}-      & \cellcolor[HTML]{E7E6E6}-      & \cellcolor[HTML]{E7E6E6}-      & \cellcolor[HTML]{E7E6E6}-      & \cellcolor[HTML]{E7E6E6}-      & \cellcolor[HTML]{E7E6E6}-      & \cellcolor[HTML]{E7E6E6}-      & \cellcolor[HTML]{E7E6E6}-      \\ \hline
                               & WEIG                         & 0.0786                         & 0.1344                         & 0.1010                         & 0.0318                         & 0.1348                         & 0.1499                         & 0.1318                         & 0.0513                         & 0.0852                         & 0.0947                         & 0.0833                         & 0.0324                         & 0.6299                         & 0.6022                         \\
                               & RoBA                         & -                              & -                              & -                              & -                              & -                              & -                              & -                              & -                              & -                              & -                              & -                              & -                              & -                              & -                              \\
\multirow{-3}{*}{4\%}          & \cellcolor[HTML]{E7E6E6}BPFA & \cellcolor[HTML]{E7E6E6}-      & \cellcolor[HTML]{E7E6E6}-      & \cellcolor[HTML]{E7E6E6}-      & \cellcolor[HTML]{E7E6E6}-      & \cellcolor[HTML]{E7E6E6}-      & \cellcolor[HTML]{E7E6E6}-      & \cellcolor[HTML]{E7E6E6}-      & \cellcolor[HTML]{E7E6E6}-      & \cellcolor[HTML]{E7E6E6}-      & \cellcolor[HTML]{E7E6E6}-      & \cellcolor[HTML]{E7E6E6}-      & \cellcolor[HTML]{E7E6E6}-      & \cellcolor[HTML]{E7E6E6}-      & \cellcolor[HTML]{E7E6E6}-      \\ \hline
                               & WEIG                         & 0.0806                         & 0.1359                         & 0.1032                         & 0.0324                         & 0.1355                         & 0.1505                         & 0.1324                         & 0.0514                         & 0.0859                         & 0.0950                         & 0.0841                         & 0.0326                         & 0.6289                         & 0.5959                         \\
                               & RoBA                         & -                              & -                              & -                              & -                              & -                              & -                              & -                              & -                              & -                              & -                              & -                              & -                              & -                              & -                              \\
\multirow{-3}{*}{7\%}          & \cellcolor[HTML]{E7E6E6}BPFA & \cellcolor[HTML]{E7E6E6}-      & \cellcolor[HTML]{E7E6E6}-      & \cellcolor[HTML]{E7E6E6}-      & \cellcolor[HTML]{E7E6E6}-      & \cellcolor[HTML]{E7E6E6}-      & \cellcolor[HTML]{E7E6E6}-      & \cellcolor[HTML]{E7E6E6}-      & \cellcolor[HTML]{E7E6E6}-      & \cellcolor[HTML]{E7E6E6}-      & \cellcolor[HTML]{E7E6E6}-      & \cellcolor[HTML]{E7E6E6}-      & \cellcolor[HTML]{E7E6E6}-      & \cellcolor[HTML]{E7E6E6}-      & \cellcolor[HTML]{E7E6E6}-      \\ \hline
                               & WEIG                         & 0.0837                         & 0.1382                         & 0.1067                         & 0.0331                         & 0.1370                         & 0.1518                         & 0.1340                         & 0.0520                         & 0.0870                         & 0.0959                         & 0.0852                         & 0.0330                         & 0.6282                         & 0.5936                         \\
                               & RoBA                         & -                              & -                              & -                              & -                              & -                              & -                              & -                              & -                              & -                              & -                              & -                              & -                              & -                              & -                              \\
\multirow{-3}{*}{10\%}         & \cellcolor[HTML]{E7E6E6}BPFA & \cellcolor[HTML]{E7E6E6}-      & \cellcolor[HTML]{E7E6E6}-      & \cellcolor[HTML]{E7E6E6}-      & \cellcolor[HTML]{E7E6E6}-      & \cellcolor[HTML]{E7E6E6}-      & \cellcolor[HTML]{E7E6E6}-      & \cellcolor[HTML]{E7E6E6}-      & \cellcolor[HTML]{E7E6E6}-      & \cellcolor[HTML]{E7E6E6}-      & \cellcolor[HTML]{E7E6E6}-      & \cellcolor[HTML]{E7E6E6}-      & \cellcolor[HTML]{E7E6E6}-      & \cellcolor[HTML]{E7E6E6}-      & \cellcolor[HTML]{E7E6E6}-      \\ \bottomrule
\end{tabular}}
\caption{Ablation study on pruning rate of PFGDFD. We use '-' to indicate that the model is completely unusable (AUC = 0.5). }
\label{tab:pruning_rate_pg}
\end{table*}

\section{Baseline Algorithm}

WEIG uses only the absolute values of the weights as the pruning score, i.e., the pruning score of WEIG is:

\begin{equation}
    PS^{WEIG}_{ijkm} = \left| W_{ijkm} \right|,
\end{equation}

RoBA uses only the reciprocal of the bias of the activations, i.e., the pruning score of RoBA is:

\begin{equation}
    PS^{RoBA}_{ijkm} = \frac{1}{BIAS_{i}}.
\end{equation}


\section{Ablation Study on Pruning Rate}

We set the pruning rates to 0.1\%, 0.4\%, 0.1\%, 1\%, 4\%, 7\%, and 10\%. At a pruning rate of 10\%. A pruning rate of 10\% is found to severely degrade utility in our experiments, so we do not conduct tests with higher pruning rates. The results are shown in the Table~\ref{tab:pruning_rate_spsl}, Table~\ref{tab:pruning_rate_ffd} and Table~\ref{tab:pruning_rate_pg}. The '-' symbol indicates that the model could completely unable to identify forged images. 
From the results, we can observe that our proposed method (BPFA) and WEIG demonstrate excellent robustness to varying pruning rates, with BPFA achieving a higher level of fairness compared to WEIG. RoBA exhibits highly unstable performance, being significantly affected by the pruning rate. Across all the pruning rates we tested, RoBA fails to achieve both good fairness and utility simultaneously. Moreover, RoBA requires extremely low pruning rates to maintain its classification capability; otherwise, it completely loses its ability to classify.


\section{Benchmark Results (Rankings)}

In Table~\ref{tab:ranking_DPD_DEOdds} and Table~\ref{tab:ranking_DEO_STD}, we present the ranking results of the benchmark, which allows for a clearer comparison of the rankings of the methods under different metrics.

\begin{table*}[ht]
\centering
\begin{tabular}{|c|c|c|c|c|c|c|}
\hline
Metric & DPD      & AADPD    & URDPD    & DEOdds   & AADEOdds & URDEOdds \\ \hline
1      & SPSL     & SPSL     & SPSL     & SPSL     & SPSL     & SPSL     \\ \hline
2      & DAW      & DAW      & DAW      & DAW      & DAW      & DAW      \\ \hline
3      & F3Net    & CORE     & CORE     & SRM      & CORE     & CORE     \\ \hline
4      & PFGDFD   & F3Net    & F3Net    & F3Net    & SRM      & SRM      \\ \hline
5      & CORE     & Capsule  & Capsule  & CORE     & F3Net    & F3Net    \\ \hline
6      & Capsule  & PFGDFD   & PFGDFD   & Capsule  & Capsule  & Capsule  \\ \hline
7      & SRM      & SRM      & SRM      & FFD      & FFD      & FFD      \\ \hline
8      & FFD      & FFD      & FFD      & RECCE    & RECCE    & RECCE    \\ \hline
9      & RECCE    & RECCE    & RECCE    & PFGDFD   & PFGDFD   & PFGDFD   \\ \hline
10     & DAG      & Xception & Xception & UCF      & UCF      & Xception \\ \hline
11     & UCF      & UCF      & DAG      & Xception & Xception & UCF      \\ \hline
12     & Xception & DA       & UCF      & DAG      & DAG      & DAG      \\ \hline
\end{tabular}
\caption{Ranking results of DPD, DEOdds and their variants. }
\label{tab:ranking_DPD_DEOdds}
\end{table*}

\begin{table*}[ht]
\centering
\begin{tabular}{|c|c|c|c|c|c|c|}
\hline
Metric & DEO      & AADEO    & URDEO    & STD      & AASTD    & URSTD    \\ \hline
1      & SPSL     & SPSL     & SPSL     & SPSL     & SPSL     & SPSL     \\ \hline
2      & DAW      & DAW      & DAW      & DAW      & DAW      & DAW      \\ \hline
3      & F3Net    & F3Net    & Capsule  & F3Net    & F3Net    & Capsule  \\ \hline
4      & PFGDFD   & CORE     & F3Net    & PFGDFD   & CORE     & F3Net    \\ \hline
5      & CORE     & PFGDFD   & CORE     & CORE     & Capsule  & CORE     \\ \hline
6      & SRM      & Capsule  & PFGDFD   & Capsule  & PFGDFD   & PFGDFD   \\ \hline
7      & Capsule  & SRM      & SRM      & SRM      & SRM      & SRM      \\ \hline
8      & FFD      & FFD      & FFD      & FFD      & FFD      & FFD      \\ \hline
9      & RECCE    & RECCE    & RECCE    & RECCE    & RECCE    & RECCE    \\ \hline
10     & UCF      & Xception & Xception & UCF      & Xception & Xception \\ \hline
11     & Xception & DAG      & DAG      & DAG      & UCF      & DAG      \\ \hline
12     & DAG      & UCF      & UCF      & Xception & DAG      & UCF      \\ \hline
\end{tabular}
\caption{Ranking results of DEO, STD and their variants. }
\label{tab:ranking_DEO_STD}
\end{table*}

\end{document}